\PassOptionsToPackage{hyphens}{url}
\PassOptionsToPackage{sort}{natbib}
\documentclass[Review,sageh,times,10pt]{sagej}
\usepackage[T1]{fontenc}
\usepackage[utf8]{inputenc}
\usepackage{graphicx,xcolor,amsmath}
\usepackage{setspace}
\usepackage{caption}
\usepackage{adjustbox}
\usepackage{etoolbox}
\renewcommand{\titlesize}{\fontsize{14}{16}\selectfont}
\apptocmd{\abstract}{\fontsize{10}{12}\selectfont}{}{}
\BeforeBeginEnvironment{threeparttable}{\begin{adjustbox}{max width=\linewidth,max totalheight=0.72\textheight}}
\AfterEndEnvironment{threeparttable}{\end{adjustbox}}
\BeforeBeginEnvironment{tabular}{\begin{adjustbox}{max width=\linewidth,max totalheight=0.72\textheight}}
\AfterEndEnvironment{tabular}{\end{adjustbox}}
\BeforeBeginEnvironment{tabularx}{\begin{adjustbox}{max width=\linewidth,max totalheight=0.72\textheight}}
\AfterEndEnvironment{tabularx}{\end{adjustbox}}
\BeforeBeginEnvironment{tikzpicture}{\begin{adjustbox}{max width=\linewidth,max totalheight=0.72\textheight}}
\AfterEndEnvironment{tikzpicture}{\end{adjustbox}}
\AtBeginEnvironment{tikzpicture}{\renewcommand{\tiny}{\fontsize{6}{7}\selectfont}\renewcommand{\scriptsize}{\fontsize{6}{7.2}\selectfont}}

\let\textcite\citet
\let\cite\citep
\usepackage[hidelinks]{hyperref}
\hypersetup{pdfauthor={},pdftitle={A Computational Framework for Modelling Organisation-Level Semantic Identity from Longitudinal Textual Data}}
\usepackage{tikz}
\usetikzlibrary{positioning,arrows.meta,fit}
\usepackage{booktabs}
\usepackage{multirow}
\usepackage{threeparttable}
\usepackage{siunitx}
\usepackage{tabularx}

\usetikzlibrary{
    positioning,
    calc,
    fit,
    calc,
    arrows.meta,
    shapes.geometric,
    shapes.multipart,
    shapes.symbols,
    shapes.misc,
    backgrounds
}

\definecolor{OSIBlue}{RGB}{52,120,246}
\definecolor{OSIGreen}{RGB}{46,170,88}
\definecolor{OSIOrange}{RGB}{236,145,33}
\definecolor{OSIPurple}{RGB}{145,85,180}
\definecolor{OSIGray}{RGB}{95,95,95}
\definecolor{OSILight}{RGB}{247,248,250}

\newcommand{\stagetitle}[1]{%
    {\bfseries\footnotesize\color{blue!50!black}#1}%
}

\newcommand{\stagecontent}[1]{%
    {\scriptsize\color{black!85}#1}%
}

\tikzset{
font=\small,
>=Latex,
flow/.style={
    -{Latex[length=3mm,width=2mm]},
    very thick,
    draw=OSIGray
},
stage/.style={
    rounded corners=3mm,
    draw=OSIGray,
    thick,
    fill=OSILight,
    minimum width=3.5cm,
    minimum height=5.2cm,
    align=center
},
centralstage/.style={
    rounded corners=3mm,
    draw=OSIGreen,
    line width=1.2pt,
    fill=green!5,
    minimum width=4.6cm,
    minimum height=6.0cm,
    align=center
},
stagetitle/.style={
    font=\bfseries\small
},
subtitle/.style={
    font=\scriptsize,
    align=center
},
concept/.style={
    circle,
    draw=OSIGreen,
    fill=green!15,
    minimum size=6mm,
    inner sep=1pt,
    font=\scriptsize
},
strongconcept/.style={
    circle,
    draw=OSIGreen,
    fill=green!30,
    minimum size=8mm,
    inner sep=1pt,
    font=\scriptsize\bfseries
},
relation/.style={
    draw=OSIGreen,
    line width=1pt
},
strongrelation/.style={
    draw=OSIGreen,
    line width=2pt
},
embedding/.style={
    circle,
    fill=OSIBlue,
    minimum size=2mm,
    inner sep=0pt
},
timeline/.style={
    draw=OSIOrange,
    line width=1pt,
    -{Latex[length=2mm]}
}
}
\begin{document}
\title{A Computational Framework for Modelling Organisation-Level Semantic Identity from Longitudinal Textual Data}
\author{Brinda Murali Krishna\affilnum{1} and Oktay Karaku\c{s}\affilnum{1} and Can Ey\"{u}po\u{g}lu\affilnum{1,2}}
\affiliation{\affilnum{1}Department of Computer Science and Informatics, Cardiff University, Cardiff, United Kingdom\\
\affilnum{2}Department of Computer Engineering, Turkish Air Force Academy, National Defence University, Istanbul, Türkiye}
\corrauth{Oktay Karaku\c{s}, Department of Computer Science and Informatics, Cardiff University, Cardiff, United Kingdom.}
\email{karakuso@cardiff.ac.uk}
\date{}

\begin{abstract}
Can longitudinal texts produced by organisationally affiliated contributors reveal collective semantic differentiation and change without obscuring differences among those contributors? We develop a computational social science framework that combines established semantic and graph measures into interpretable organisational fingerprints and temporal profiles. We use \emph{organisation-level semantic identity} to describe a corpus-dependent interpretation of recurring and changing features of affiliated textual output, rather than corporate culture or managerial intention. Drawing on an inherited semantic landscape, we examine 1,289 songs by 126 artists affiliated with four K-pop entertainment companies between 1997 and 2023. The evidence for differentiation is selective. After correction, all three novelty-related measures show artist-level differences, whereas the diversity and connectivity measures tested do not; only two company--community enrichments remain significant. Temporal trajectories reveal further distinctions between portfolios. Artist-grouped balanced accuracy reaches 0.334 against a chance baseline of 0.25, providing modest supporting evidence of affiliation-related signal rather than independent validation of identity. Robustness analyses support the presence of continuous semantic structure more strongly than precise rankings. Together, these findings show how collective regularities can coexist with shared content, contributor heterogeneity and change over time, while leaving cross-domain generality and broader organisational construct validity unresolved.
\end{abstract}

\keywords{computational social science, organisational semantic identity, longitudinal text analysis, semantic fingerprints, cultural production, K-pop}
\maketitle
\section{Introduction}
\label{sec:introduction}
Can longitudinal texts produced by organisationally affiliated contributors reveal collective semantic differentiation and change without obscuring differences among those contributors? The question becomes particularly important when the evidence consists of cultural works released by affiliated artists rather than statements issued by an organisation itself. Patterns visible at portfolio level may reflect recurring features of the collective output, yet they can coexist with considerable variation among artists and substantial overlap between companies.

Computational research has used organisational text to investigate culture, communication and positioning \cite{Li2021CorporateCulture,Sull2019Culture500,Koch2023CultureBERT,Schachner2024CultureDictionary}. More recent work in computational social science brings textual and network methods to specific questions about business brand communication and political-party communication \cite{Su2024HumanInsights,Swalve2025PartyCommunication}. Applying related methods to affiliated cultural production raises a distinct inferential challenge: grouping works by company does not establish that the company authored, directed or controlled their content.

To address this challenge, we distinguish three analytical objects. A semantic representation provides the computational coordinates in which texts can be compared, while a semantic fingerprint summarises measurable characteristics across several dimensions. By \emph{organisation-level semantic identity}, we mean a corpus-dependent interpretation of recurring and evolving patterns in those fingerprints, assessed through comparative and temporal evidence. This is not a learned identity estimator, nor does it directly measure corporate culture, members' self-understanding, managerial intention or strategy. Organisational identity theory motivates attention to distinctiveness and continuity alongside change \cite{Albert1985,Gioia2000AdaptiveInstability}, but establishing the validity of a broader organisational construct requires evidence beyond computational performance \cite{Joung2026Measurement}.

Our empirical setting comprises 1,289 songs by 126 artists affiliated with four K-pop entertainment companies over 1997--2023. Building on an inherited semantic landscape \cite{karakucs2026semantic}, we examine community composition, diversity, concentration, connectivity and novelty, connecting portfolio-level descriptions with artist-level comparisons and longitudinal profiles. Rather than introducing a new embedding model or graph algorithm, the study brings established measures together to address a bounded social-science question about collective patterns in affiliated cultural output.

The resulting picture is neither one of uniformly distinct companies nor one of complete semantic convergence. Artist-level differences remain significant after correction for novelty-related measures, but not for the diversity and connectivity measures tested, and only two company--community enrichments survive correction. Temporal trajectories add qualifications that aggregate fingerprints alone cannot capture. Modest artist-grouped predictive performance provides complementary evidence of affiliation-related signal, while robustness, catalogue recovery and persistence analyses assess properties of the representation rather than independently validating organisational identity.

The study therefore contributes a conceptual distinction between measurement and interpretation, a computational approach that retains contributor heterogeneity within portfolio comparisons, and a longitudinal account of continuity and change. Empirically, it identifies supported differences alongside shared patterns and null findings. These conclusions concern four organisations within one cultural domain; applying the framework elsewhere would require renewed assessment of both the semantic representation and the proposed construct. Sections~\ref{sec:related_work}--\ref{sec:validation} introduce the relevant literature, empirical setting and methods, followed by the findings and their implications in Sections~\ref{sec:results}--\ref{sec:conclusion}.

\section{Related Work}
\label{sec:related_work}
\subsection{Organisational Text and Interpretable Comparison}
\label{sec:organisation_text_analysis}
\label{sec:organisation_semantic_identity}
Company embeddings represent organisational text in ways that support similarity assessment, retrieval and prediction \cite{Ito2020CompanyEmbeddings,Gerling2024Company2Vec,Vamvourellis2023CompanySimilarity,Molinari2024SparseCompanySimilarity}. Multimodal approaches incorporate additional sources of organisational information \cite{Dolphin2023MultimodalIndustry}. Although valuable for these tasks, proximity in a latent space does not, by itself, explain which characteristics shape a comparison, how contributors vary within an organisation or how those characteristics develop over time. A multidimensional fingerprint makes these properties separately observable, leaving semantic identity as an interpretation of their combined pattern rather than a property directly read from an embedding.

A related body of computational culture research measures specified constructs using organisational texts, employee reviews, transformer models or dictionaries \cite{Li2021CorporateCulture,Sull2019Culture500,Koch2023CultureBERT,Schachner2024CultureDictionary}. Studies of communicated identity likewise examine how organisations position themselves through language \cite{Toschi2023SocialImpactIdentity}. This literature demonstrates the value of interpretable operationalisation, but neither its underlying constructs nor its textual sources are interchangeable with lyrics produced by affiliated artists.

\subsection{Semantic Structure and Longitudinal Comparison}
\label{sec:semantic_representation_learning}
\label{sec:temporal_semantic_modelling}
Contextual language representations and embedding-based topic models offer ways to locate documents within shared semantic spaces and identify recurring groupings \cite{Devlin2019,Reimers2019,Dieng2020ETM,Grootendorst2022BERTopic}. Graph representation learning provides complementary tools for studying relationships among observations \cite{Hamilton2020GRL}. In this study, however, the multilingual embeddings, similarity graph and communities come from an earlier analysis of lyrics \cite{karakucs2026semantic}. We treat them as fixed analytical resources rather than claiming their construction as a new methodological contribution.

Dynamic embeddings and topic models have been developed to examine how linguistic structure evolves \cite{Bamler2017DynamicEmbeddings,Rudolph2018DynamicEmbeddings,Blei2006DynamicTopicModels}. Our approach asks a different temporal question: how do organisational portfolios change when compared across rolling windows in a fixed semantic coordinate system? Holding that system constant allows community composition to be compared across periods, although the resulting trajectories necessarily remain conditional on the inherited representation.

\subsection{Identity, Culture and the Inferential Boundary}
\label{sec:organisational_identity}
Organisational identity has been associated with characteristics understood as central, enduring and distinctive \cite{Albert1985}, while accounts of adaptive instability emphasise that continuity can coexist with change \cite{Gioia2000AdaptiveInstability}. Organisational, corporate and social identity are related but place different weight on members' understandings and external expression; organisational culture, meanwhile, concerns shared values and practices \cite{Cornelissen2007IdentityIntegration,Dowling2011CorporateOrganizationalIdentity}. Similarities among lyrics cannot establish any of these constructs directly, making it important to separate the patterns we measure from the meanings we might attribute to them.

Recent studies in \emph{Social Science Computer Review} illustrate how computational methods can address carefully delimited organisational questions. Human-informed topic modelling has been applied to businesses' social-media brand communication \cite{Su2024HumanInsights}, while network analysis has examined centralisation in political-party communication with attention to differences between participant roles \cite{Swalve2025PartyCommunication}. These studies offer useful examples of question-led computational analysis, not evidence that lyrics measure corporate culture. Work on computational measurement also reinforces the distinction between technical performance and external construct validity \cite{Joung2026Measurement}.

\subsection{Positioning}
\label{sec:positioning_framework}
Against this background, we combine a shared semantic representation with multidimensional fingerprints, contributor-level comparisons and longitudinal evidence to examine organisationally grouped cultural output. Distinguishing representation, fingerprint and identity keeps the interpretation anchored to observable measures while making room for null findings and sensitivity to analytical choices. The framework is intended to clarify where collective differentiation is supported and where overlap or uncertainty constrains interpretation; its applicability beyond this domain, and its correspondence with broader organisational constructs, remain to be established.

\section{Problem Formulation and Empirical Setting}
\label{sec:problem}
\subsection{Analytical Setting and Data Hierarchy}
Let $c\in\mathcal{C}$ denote a company, $a\in\mathcal{A}_c$ an assigned artist, $s\in\mathcal{S}_a$ a song, $\ell_s$ its lyrics, and $t$ its release year. The analytical hierarchy is
\[
\ell_s
\rightarrow
s
\rightarrow
a
\rightarrow
c
\leftrightarrow
t.
\]
In this hierarchy, songs provide semantic observations, artists serve as the units of inference, and companies are described through portfolio summaries that can also be compared over time. Affiliation determines how observations are grouped, but does not establish authorship, organisational direction or intention.

\subsection{Organisational Semantic Identity and Objective}
\textbf{Definition 1 (Organisational Semantic Identity).} \emph{Within this study, organisational semantic identity is a corpus-dependent computational construct: a qualified interpretation of recurring and evolving semantic characteristics in a company-associated lyrical portfolio, derived from measurable fingerprints and comparative, temporal and evidential qualifications.}

The fingerprints are quantitative descriptions, whereas identity refers to their qualified interpretation. We therefore examine where portfolios differ, how their characteristics persist or change, and what the available evidence can support. Statistical significance is evaluated rather than assumed to be necessary across every dimension; shared patterns and differences lacking statistical support are also informative.

\subsection{Corpus Provenance and Company Attribution}
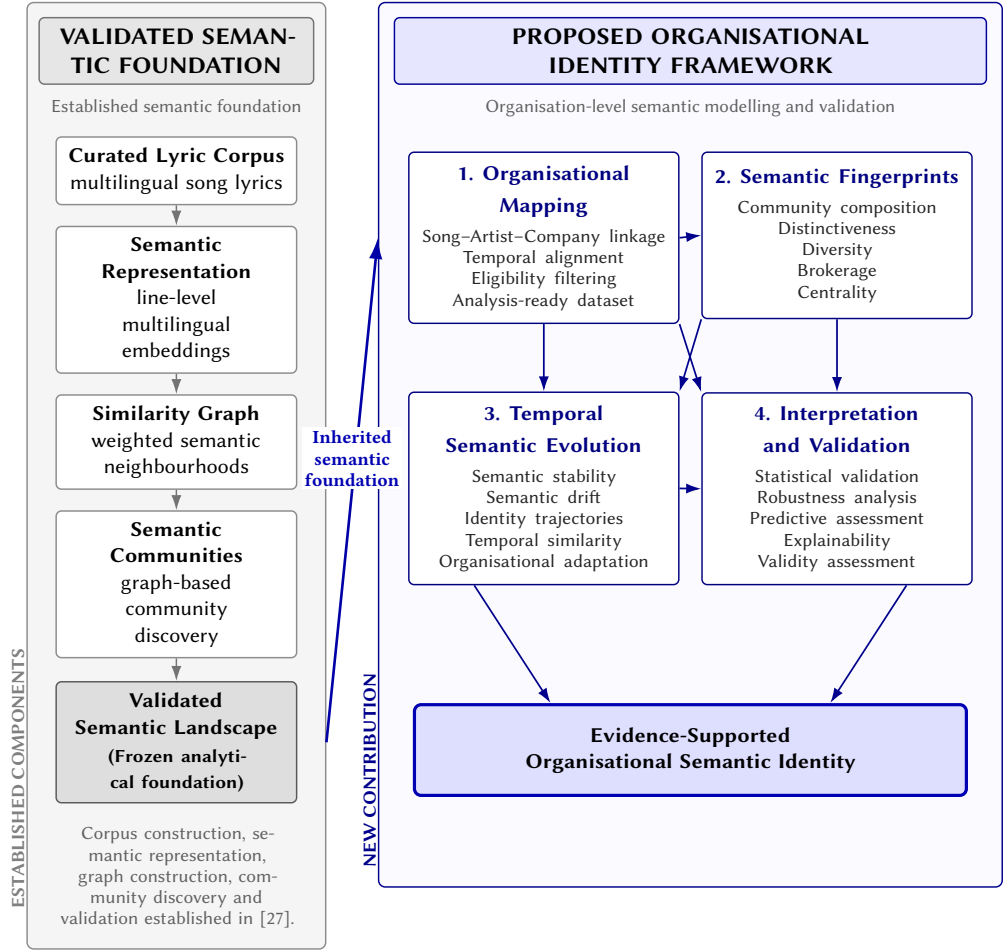
\begin{figure}[t]
\centering
\begingroup
\def\small{\fontsize{9}{10.8}\selectfont}
\def\footnotesize{\fontsize{8}{9.6}\selectfont}
\def\scriptsize{\fontsize{7}{8.4}\selectfont}
\fontsize{10}{12}\selectfont
\begin{tikzpicture}[
    font=\sffamily,
    >=Latex,
    line width=0.55pt,
    every node/.style={
        align=center
    },
    previouspanel/.style={
        draw=black!45,
        rounded corners=2.5pt,
        fill=black!4,
        minimum width=3.95cm,
        minimum height=12.52cm,
        inner sep=0pt
    },
    currentpanel/.style={
        draw=blue!55!black,
        rounded corners=2.5pt,
        fill=blue!2,
        minimum width=8.25cm,
        minimum height=11.70cm,
        inner sep=0pt
    },
    previousheader/.style={
        draw=black!55,
        rounded corners=2pt,
        fill=black!10,
        text width=3.35cm,
        minimum height=0.72cm,
        font=\sffamily\bfseries\small,
        inner sep=4pt
    },
    currentheader/.style={
        draw=blue!60!black,
        rounded corners=2pt,
        fill=blue!10,
        text width=7.55cm,
        minimum height=0.72cm,
        font=\sffamily\bfseries\small,
        inner sep=4pt
    },
    prevnode/.style={
        draw=black!45,
        rounded corners=2pt,
        fill=white,
        text width=2.90cm,
        minimum height=0.68cm,
        font=\sffamily\footnotesize,
        inner sep=4pt
    },
    prevfinal/.style={
        draw=black!65,
        rounded corners=2pt,
        fill=black!13,
        text width=2.90cm,
        minimum height=0.84cm,
        font=\sffamily\bfseries\footnotesize,
        inner sep=4pt
    },
    stagebox/.style={
        draw=blue!50!black,
        rounded corners=2pt,
        fill=white,
        text width=3.22cm,
        minimum height=2.20cm,
        inner sep=5pt
    },
    identitybox/.style={
        draw=blue!70!black,
        line width=1.2pt,
        rounded corners=2.5pt,
        fill=blue!13,
        text width=6.95cm,
        minimum height=1.20cm,
        font=\sffamily\bfseries\footnotesize,
        inner sep=5pt
    },
    smalllabel/.style={
        font=\sffamily\scriptsize,
        text=black!62
    },
    flow/.style={
        -{Latex[length=2.0mm,width=1.35mm]},
        draw=black!55,
        line width=0.55pt
    },
    currentflow/.style={
        -{Latex[length=2.0mm,width=1.35mm]},
        draw=blue!55!black,
        line width=0.65pt
    },
    bridge/.style={
        -{Latex[length=2.8mm,width=1.8mm]},
        draw=blue!65!black,
        line width=1.0pt
    }
]
\node[previouspanel, anchor=north west] (previous-panel)
    at (0,0) {};
\node[currentpanel, anchor=north west] (current-panel)
    at (4.65,0) {};
\node[previousheader, anchor=north]
    (previous-header)
    at ([yshift=-0.20cm]previous-panel.north)
    {VALIDATED SEMANTIC FOUNDATION};
\node[smalllabel, below=0.06cm of previous-header]
    {Established semantic foundation};
\node[prevnode, below=0.68cm of previous-header.south]
    (lyrics)
    {\textbf{Curated Lyric Corpus}\\
     multilingual song lyrics};
\node[prevnode, below=0.27cm of lyrics]
    (representation)
    {\textbf{Semantic\\Representation}\\
     line-level\\multilingual\\embeddings};
\node[prevnode, below=0.27cm of representation]
    (graph)
    {\textbf{Similarity Graph}\\
     weighted semantic neighbourhoods};
\node[prevnode, below=0.27cm of graph]
    (communities)
    {\textbf{Semantic Communities}\\
     graph-based\\community\\discovery};
\node[prevfinal, below=0.30cm of communities]
    (landscape)
    {\textbf{Validated}\\
\textbf{Semantic Landscape}\\[0.2mm]
\scriptsize
(Frozen analytical foundation)};
\draw[flow] (lyrics) -- (representation);
\draw[flow] (representation) -- (graph);
\draw[flow] (graph) -- (communities);
\draw[flow] (communities) -- (landscape);
\node[
    smalllabel,
    text width=3.15cm,
    anchor=south
]
    at ([yshift=0.16cm]previous-panel.south)
    {Corpus construction, semantic representation,
     graph construction, community discovery and validation
     established in~\cite{karakucs2026semantic}.};
\node[currentheader, anchor=north]
    (current-header)
    at ([yshift=-0.20cm]current-panel.north)
    {PROPOSED ORGANISATIONAL\\IDENTITY FRAMEWORK};
\node[smalllabel, below=0.06cm of current-header]
    {Organisation-level semantic modelling and validation};
\node[
    stagebox,
    anchor=north west
]
(mapping)
at ([xshift=0.40cm,yshift=-1.98cm]current-panel.north west)
{
    \stagetitle{1. Organisational\\Mapping}\\[0.32em]
    \stagecontent{
Song--Artist--Company linkage\\
Temporal alignment\\
Eligibility filtering\\
Analysis-ready dataset\\
}
};

\node[
    stagebox,
    anchor=north east
]
(fingerprints)
at ([xshift=-0.40cm,yshift=-1.98cm]current-panel.north east)
{
    \stagetitle{2. Semantic Fingerprints}\\[0.32em]
    \stagecontent{
Community composition\\
Distinctiveness\\
Diversity\\
Brokerage\\
Centrality\\
}};

\node[
    stagebox,
    anchor=south west
]
(temporal)
at ([xshift=0.40cm,yshift=4.0cm]current-panel.south west)
{
    \stagetitle{3. Temporal\\Semantic Evolution}\\[0.32em]
    \stagecontent{
Semantic stability\\
Semantic drift\\
Identity trajectories\\
Temporal similarity\\
Organisational adaptation\\
}
};

\node[
    stagebox,
    anchor=south east
]
(validation)
at ([xshift=-0.40cm,yshift=4.0cm]current-panel.south east)
{
    \stagetitle{4. Interpretation and Validation}\\[0.32em]
    \stagecontent{
Statistical validation\\
Robustness analysis\\
Predictive assessment\\
Explainability\\
Validity assessment\\
}
};
\draw[currentflow]
    (mapping.east) -- (fingerprints.west);
\draw[currentflow]
    (mapping.south) -- (temporal.north);
\draw[currentflow]
    (fingerprints.south) -- (validation.north);
\draw[currentflow]
    (temporal.east) -- (validation.west);
\draw[currentflow]
(fingerprints.south west)
--
(temporal.north east);
\draw[currentflow]
(mapping.south east)
--
(validation.north west);
\node[
    identitybox,
    anchor=south
]
    (identity)
    at ([yshift=1.20cm]current-panel.south)
    {Evidence-Supported\\
Organisational Semantic Identity};
\draw[currentflow]
    ([xshift=-0.92cm]temporal.south) -- ([xshift=-1.85cm]identity.north);
\draw[currentflow]
    ([xshift=0.92cm]validation.south) -- ([xshift=1.85cm]identity.north);
\coordinate (bridge-start)
    at ([xshift=0.00cm]previous-panel.east |- landscape.east);
\coordinate (bridge-end)
    at ([xshift=-0.00cm]current-panel.west |- mapping.west);
\draw[bridge]
    (bridge-start)
    -- node[
    above,
    font=\bfseries\scriptsize,
    align=center,
    fill=white,
    inner sep=1pt,
    text=blue!70!black
]
{
Inherited\\
semantic\\
foundation
}
    (bridge-end);
\node[
    rotate=90,
    font=\sffamily\scriptsize\bfseries,
    text=black!50,
    anchor=south
]
    at ([xshift=0.10cm,yshift=-4.0cm]previous-panel.west)
    {ESTABLISHED COMPONENTS};
\node[
    rotate=90,
    font=\sffamily\scriptsize\bfseries,
    text=blue!55!black,
    anchor=south
]
    at ([xshift=0.10cm,yshift=-4.3cm]current-panel.west)
    {NEW CONTRIBUTION};

\end{tikzpicture}
\endgroup
\caption{
Relationship between the validated semantic foundation established in ~\cite{karakucs2026semantic} and the organisational semantic identity framework proposed in this paper. The study in~\cite{karakucs2026semantic} constructs a multilingual semantic landscape through representation learning, graph construction, community discovery and semantic validation. Building upon this frozen analytical foundation, the present work introduces an organisation-level framework that models entertainment companies through hierarchical aggregation, multidimensional semantic fingerprints, temporal semantic evolution and comprehensive validation. The contribution boundary separates the inherited semantic landscape from the present integration of established measures for organisation-level analysis. Identity denotes a corpus-dependent interpretation of the fingerprints; the diagram does not imply a new identity-learning algorithm or direct measurement of corporate identity.
}
\label{fig:framework_relationship}
\end{figure}

The corpus, preprocessing, duplicate handling, multilingual representation, similarity graph and validated communities originate in \textcite{karakucs2026semantic}. Figure~\ref{fig:framework_relationship} separates this inherited foundation from the present organisational analysis.
The retained data-construction code normalises artist names and assigns companies through a static lookup, recording its basis as \texttt{static\_student\_lookup} and indicating that company affiliation was not verified at each release date. Ambiguous name matches are flagged and excluded from the default eligible dataset. Because the lookup does not reconstruct historical label changes, subsidiary relationships or acquisitions, company labels represent fixed analytical portfolios rather than verified ownership at the time of release. Temporal comparisons consequently describe changes within those assigned portfolios.

\subsection{Eligibility and Dataset Characteristics}
Only records satisfying
\[
\texttt{analysis\_eligible}=\texttt{True}
\]
are retained for analysis. The analytical records provide lyrics-derived representations, artist and company identifiers, release years, community assignments and downstream metrics. The dataset contains 1,289 songs by 126 artists assigned to four companies during 1997--2023 and spans fifteen inherited communities (Table~\ref{tab:dataset_summary}). The dataset supports comparisons among these observed portfolios, but not representative inference about entertainment organisations more generally.
\begin{table}[t]
\centering
\caption{Summary of the analysed longitudinal K-pop lyric dataset after application of the eligibility criteria.}
\label{tab:dataset_summary}
\begin{tabular}{lccccc}
\toprule
Company & Songs & Artists & First Year & Last Year & Communities\\
\midrule
HYBE & 170 & 15 & 2003 & 2023 & 14\\
JYP & 178 & 19 & 2001 & 2023 & 15\\
SM & 528 & 59 & 1998 & 2023 & 15\\
YG & 413 & 33 & 1997 & 2023 & 15\\
\midrule
Total & 1,289 & 126 & 1997 & 2023 & 15\\
\bottomrule
\end{tabular}
\end{table}

\section{Computational Framework for Organisational Semantic Identity}
\label{sec:framework}
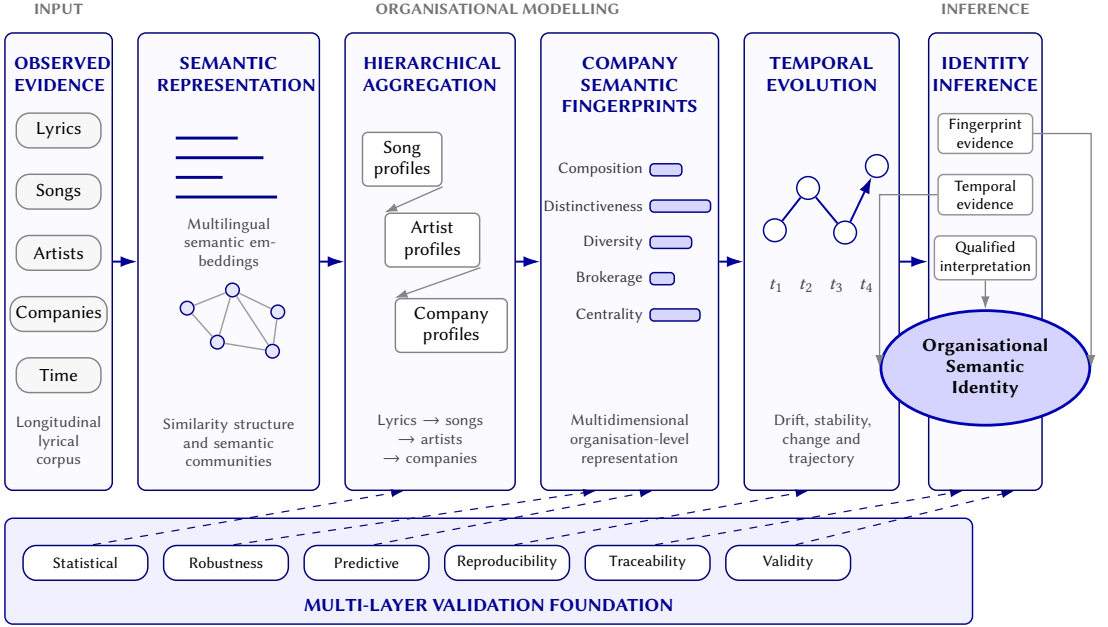
\begin{figure}[t]
\centering

\begingroup
\def\small{\fontsize{9}{10.8}\selectfont}
\def\footnotesize{\fontsize{8}{9.6}\selectfont}
\def\scriptsize{\fontsize{7}{8.4}\selectfont}
\fontsize{10}{12}\selectfont
\begin{tikzpicture}[
    x=1cm,
    y=1cm,
    font=\sffamily,
    >=Latex,
    line width=0.55pt,
    every node/.style={
        align=center
    },
    stagearea/.style={
        draw=blue!45!black,
        rounded corners=3pt,
        fill=blue!2,
        inner sep=5pt
    },
    stageheading/.style={
        font=\sffamily\bfseries\scriptsize,
        text=blue!55!black
    },
    stagesubtext/.style={
        font=\sffamily\tiny,
        text=black!72
    },
    inputtoken/.style={
        draw=black!48,
        rounded corners=5pt,
        fill=black!3,
        minimum width=1.12cm,
        minimum height=0.46cm,
        font=\sffamily\scriptsize,
        inner sep=2pt
    },
    semanticnode/.style={
        circle,
        draw=blue!50!black,
        fill=blue!10,
        minimum size=0.18cm,
        inner sep=0pt
    },
    communitynode/.style={
        circle,
        draw=blue!55!black,
        fill=blue!18,
        minimum size=0.27cm,
        inner sep=0pt
    },
    aggregation/.style={
        draw=black!48,
        rounded corners=2pt,
        fill=white,
        minimum height=0.51cm,
        font=\sffamily\scriptsize,
        inner sep=3pt
    },
    fingerprintlabel/.style={
        anchor=east,
        font=\sffamily\tiny,
        text=black!78
    },
    fingerprintbar/.style={
        rounded corners=1.5pt,
        draw=blue!50!black,
        fill=blue!16,
        minimum height=0.16cm,
        anchor=west,
        inner sep=0pt
    },
    temporalstate/.style={
        circle,
        draw=blue!55!black,
        fill=white,
        minimum size=0.30cm,
        inner sep=0pt
    },
    evidenceitem/.style={
        draw=black!42,
        rounded corners=2pt,
        fill=white,
        minimum width=1.35cm,
        minimum height=0.42cm,
        font=\sffamily\tiny,
        inner sep=2pt
    },
    identityoutput/.style={
        ellipse,
        draw=blue!70!black,
        line width=1.05pt,
        fill=blue!17,
        minimum width=2.10cm,
        minimum height=1.50cm,
        font=\sffamily\bfseries\scriptsize,
        inner sep=4pt
    },
    validationband/.style={
        draw=blue!60!black,
        rounded corners=3pt,
        fill=blue!6,
        minimum width=12.80cm,
        minimum height=1.40cm,
        inner sep=5pt
    },
    validationitem/.style={
        draw=blue!45!black,
        rounded corners=5pt,
        fill=white,
        minimum width=1.65cm,
        minimum height=0.46cm,
        font=\sffamily\tiny,
        inner sep=2pt
    },
    layerlabel/.style={
        font=\sffamily\bfseries\tiny,
        text=black!55
    },
    mainflow/.style={
        -{Latex[length=2.2mm,width=1.45mm]},
        draw=blue!60!black,
        line width=0.75pt
    },
    secondaryflow/.style={
        -{Latex[length=1.7mm,width=1.15mm]},
        draw=black!48,
        line width=0.50pt
    },
    supportflow/.style={
        -{Latex[length=1.65mm,width=1.10mm]},
        draw=blue!48!black,
        line width=0.45pt,
        dashed
    }
]


\node[
    stagearea,
    minimum width=1.42cm,
    minimum height=6.05cm,
    anchor=north west
]
    (input-area)
    at (0,0)
    {};

\node[
    stageheading,
    text width=1.18cm,
    anchor=north
]
    at ([yshift=-0.20cm]input-area.north)
    {OBSERVED\\EVIDENCE};

\node[inputtoken]
    (lyrics)
    at ([yshift=-1.30cm]input-area.north)
    {Lyrics};

\node[inputtoken, below=0.33cm of lyrics]
    (songs)
    {Songs};

\node[inputtoken, below=0.33cm of songs]
    (artists)
    {Artists};

\node[inputtoken, below=0.33cm of artists]
    (companies)
    {Companies};

\node[inputtoken, below=0.33cm of companies]
    (time)
    {Time};

\node[
    stagesubtext,
    text width=1.15cm,
    anchor=south
]
    at ([yshift=0.18cm]input-area.south)
    {Longitudinal\\lyrical corpus};


\node[
    stagearea,
    minimum width=2.40cm,
    minimum height=6.05cm,
    anchor=north west
]
    (semantic-area)
    at (1.76,0)
    {};

\node[
    stageheading,
    text width=1.90cm,
    anchor=north
]
    at ([yshift=-0.20cm]semantic-area.north)
    {SEMANTIC\\REPRESENTATION};

\foreach \yy/\ww in {-1.40/0.82,-1.66/1.16,-1.92/0.62,-2.18/1.34} {
    \draw[
        blue!55!black,
        line width=1.10pt,
        rounded corners=1pt
    ]
    ([xshift=-0.70cm,yshift=\yy cm]semantic-area.north)
    --
    ++(\ww,0);
}

\node[
    stagesubtext,
    text width=1.75cm
]
    at ([yshift=-2.80cm]semantic-area.north)
    {Multilingual\\semantic embeddings};

\node[semanticnode]
    (sg1)
    at ([xshift=-0.55cm,yshift=-3.65cm]semantic-area.north)
    {};

\node[semanticnode]
    (sg2)
    at ([xshift=0.05cm,yshift=-3.41cm]semantic-area.north)
    {};

\node[semanticnode]
    (sg3)
    at ([xshift=0.65cm,yshift=-3.70cm]semantic-area.north)
    {};

\node[semanticnode]
    (sg4)
    at ([xshift=-0.15cm,yshift=-4.10cm]semantic-area.north)
    {};

\node[semanticnode]
    (sg5)
    at ([xshift=0.58cm,yshift=-4.22cm]semantic-area.north)
    {};

\draw[draw=black!38] (sg1) -- (sg2);
\draw[draw=black!38] (sg2) -- (sg3);
\draw[draw=black!38] (sg1) -- (sg4);
\draw[draw=black!38] (sg2) -- (sg4);
\draw[draw=black!38] (sg2) -- (sg5);
\draw[draw=black!38] (sg3) -- (sg5);
\draw[draw=black!38] (sg4) -- (sg5);

\node[
    stagesubtext,
    text width=1.75cm,
    anchor=south
]
    at ([yshift=0.20cm]semantic-area.south)
    {Similarity structure\\and semantic communities};


\node[
    stagearea,
    minimum width=2.25cm,
    minimum height=6.05cm,
    anchor=north west
]
    (aggregation-area)
    at (4.50,0)
    {};

\node[
    stageheading,
    text width=1.76cm,
    anchor=north
]
    at ([yshift=-0.20cm]aggregation-area.north)
    {HIERARCHICAL\\AGGREGATION};

\node[
    aggregation,
    minimum width=1.05cm
]
    (song-profile)
    at ([xshift=-0.37cm,yshift=-1.68cm]aggregation-area.north)
    {Song\\profiles};

\node[
    aggregation,
    minimum width=1.25cm
]
    (artist-profile)
    at ([xshift=0.03cm,yshift=-2.75cm]aggregation-area.north)
    {Artist\\profiles};

\node[
    aggregation,
    minimum width=1.48cm
]
    (company-profile)
    at ([xshift=0.28cm,yshift=-3.88cm]aggregation-area.north)
    {Company\\profiles};

\draw[secondaryflow]
    (song-profile.south east)
    --
    (artist-profile.north west);

\draw[secondaryflow]
    (artist-profile.south east)
    --
    (company-profile.north west);

\node[
    stagesubtext,
    text width=1.72cm,
    anchor=south
]
    at ([yshift=0.20cm]aggregation-area.south)
    {Lyrics $\rightarrow$ songs\\
     $\rightarrow$ artists\\
     $\rightarrow$ companies};


\node[
    stagearea,
    minimum width=2.35cm,
    minimum height=6.05cm,
    anchor=north west
]
    (fingerprint-area)
    at (7.09,0)
    {};

\node[
    stageheading,
    text width=2.02cm,
    anchor=north
]
    at ([yshift=-0.20cm]fingerprint-area.north)
    {COMPANY SEMANTIC\\FINGERPRINTS};

\node[fingerprintlabel]
    at ([xshift=0.30cm,yshift=-1.82cm]fingerprint-area.north)
    {Composition};

\node[
    fingerprintbar,
    minimum width=0.42cm
]
    at ([xshift=0.26cm,yshift=-1.82cm]fingerprint-area.north)
    {};

\node[fingerprintlabel]
    at ([xshift=0.3cm,yshift=-2.30cm]fingerprint-area.north)
    {Distinctiveness};

\node[
    fingerprintbar,
    minimum width=0.80cm
]
    at ([xshift=0.26cm,yshift=-2.30cm]fingerprint-area.north)
    {};

\node[fingerprintlabel]
    at ([xshift=0.30cm,yshift=-2.78cm]fingerprint-area.north)
    {Diversity};

\node[
    fingerprintbar,
    minimum width=0.55cm
]
    at ([xshift=0.26cm,yshift=-2.78cm]fingerprint-area.north)
    {};

\node[fingerprintlabel]
    at ([xshift=0.30cm,yshift=-3.26cm]fingerprint-area.north)
    {Brokerage};

\node[
    fingerprintbar,
    minimum width=0.32cm
]
    at ([xshift=0.26cm,yshift=-3.26cm]fingerprint-area.north)
    {};

\node[fingerprintlabel]
    at ([xshift=0.30cm,yshift=-3.74cm]fingerprint-area.north)
    {Centrality};

\node[
    fingerprintbar,
    minimum width=0.66cm
]
    at ([xshift=0.26cm,yshift=-3.74cm]fingerprint-area.north)
    {};

\node[
    stagesubtext,
    text width=2.00cm,
    anchor=south
]
    at ([yshift=0.20cm]fingerprint-area.south)
    {Multidimensional\\organisation-level\\representation};


\node[
    stagearea,
    minimum width=2.05cm,
    minimum height=6.05cm,
    anchor=north west
]
    (temporal-area)
    at (9.78,0)
    {};

\node[
    stageheading,
    text width=1.73cm,
    anchor=north
]
    at ([yshift=-0.20cm]temporal-area.north)
    {TEMPORAL\\EVOLUTION};

\node[temporalstate]
    (t1)
    at ([xshift=-0.62cm,yshift=-2.62cm]temporal-area.north)
    {};

\node[temporalstate]
    (t2)
    at ([xshift=-0.18cm,yshift=-2.06cm]temporal-area.north)
    {};

\node[temporalstate]
    (t3)
    at ([xshift=0.31cm,yshift=-2.65cm]temporal-area.north)
    {};

\node[temporalstate]
    (t4)
    at ([xshift=0.73cm,yshift=-1.77cm]temporal-area.north)
    {};

\draw[
    mainflow,
    rounded corners=4pt
]
    (t1)
    --
    (t2)
    --
    (t3)
    --
    (t4);

\node[
    stagesubtext
]
    at ([yshift=-3.35cm]temporal-area.north)
    {$t_1 \quad t_2 \quad t_3 \quad t_4$};

\node[
    stagesubtext,
    text width=1.72cm,
    anchor=south
]
    at ([yshift=0.20cm]temporal-area.south)
    {Drift, stability,\\change and\\trajectory};


\node[
    stagearea,
    minimum width=1.5cm,
    minimum height=6.05cm,
    anchor=north west
]
    (identity-area)
    at (12.22,0)
    {};

\node[
    stageheading,
    text width=1.58cm,
    anchor=north
]
    at ([yshift=-0.20cm]identity-area.north)
    {IDENTITY\\INFERENCE};

\node[
    evidenceitem,
    minimum width=1.25cm
]
    (quant-evidence)
    at ([yshift=-1.34cm]identity-area.north)
    {Fingerprint\\evidence};

\node[
    evidenceitem,
    minimum width=1.25cm,
    below=0.26cm of quant-evidence
]
    (temporal-evidence)
    {Temporal\\evidence};

\node[
    evidenceitem,
    minimum width=1.25cm,
    below=0.26cm of temporal-evidence
]
    (interpretation)
    {Qualified\\interpretation};

\node[
    identityoutput,
    minimum width=1.56cm,
    minimum height=1.18cm,
    below=0.38cm of interpretation
]
    (identity)
    {Organisational\\Semantic\\Identity};

\draw[secondaryflow]
    (quant-evidence.east)
    -|
    (identity.east);

\draw[secondaryflow]
    (temporal-evidence.west)
    -|
    (identity.west);

\draw[secondaryflow]
    (interpretation)
    --
    (identity);


\draw[mainflow]
    (input-area.east)
    --
    (semantic-area.west);

\draw[mainflow]
    (semantic-area.east)
    --
    (aggregation-area.west);

\draw[mainflow]
    (aggregation-area.east)
    --
    (fingerprint-area.west);

\draw[mainflow]
    (fingerprint-area.east)
    --
    (temporal-area.west);

\draw[mainflow]
    (temporal-area.east)
    --
    (identity-area.west);


\node[
    validationband,
    anchor=north west
]
    (validation-band)
    at (0,-6.42)
    {};

\node[
    stageheading,
    anchor=north
]
    at ([yshift=-0.95cm]validation-band.north)
    {MULTI-LAYER VALIDATION FOUNDATION};

\node[
    validationitem,
    anchor=west
]
    (statistical)
    at ([xshift=0.24cm,yshift=-0.6cm]validation-band.north west)
    {Statistical};

\node[
    validationitem,
    right=0.19cm of statistical
]
    (robustness)
    {Robustness};

\node[
    validationitem,
    right=0.19cm of robustness
]
    (predictive)
    {Predictive};

\node[
    validationitem,
    right=0.19cm of predictive
]
    (reproducibility)
    {Reproducibility};

\node[
    validationitem,
    right=0.19cm of reproducibility
]
    (traceability)
    {Traceability};

\node[
    validationitem,
    right=0.19cm of traceability
]
    (validity)
    {Validity};

\draw[supportflow]
    ([xshift=0.10cm]statistical.north)
    --
    ([xshift=-0.35cm]aggregation-area.south);

\draw[supportflow]
    ([xshift=0.10cm]robustness.north)
    --
    ([xshift=-0.45cm]fingerprint-area.south);

\draw[supportflow]
    ([xshift=0.10cm]predictive.north)
    --
    ([xshift=0.30cm]fingerprint-area.south);

\draw[supportflow]
    ([xshift=0.10cm]reproducibility.north)
    --
    ([xshift=-0.18cm]temporal-area.south);

\draw[supportflow]
    ([xshift=0.10cm]traceability.north)
    --
    ([xshift=-0.30cm]identity-area.south);

\draw[supportflow]
    ([xshift=0.10cm]validity.north)
    --
    ([xshift=0.38cm]identity-area.south);


\node[
    layerlabel,
    anchor=south
]
    at ([yshift=0.11cm]input-area.north)
    {INPUT};

\node[
    layerlabel,
    anchor=south
]
    at ([xshift=-1.75cm,yshift=0.11cm]fingerprint-area.north)
    {ORGANISATIONAL MODELLING};

\node[
    layerlabel,
    anchor=south
]
    at ([yshift=0.11cm]identity-area.north)
    {INFERENCE};

\end{tikzpicture}
\endgroup

\caption{
End-to-end computational framework for organisational semantic identity
modelling. Longitudinal lyrical observations and associated artist, company
and temporal metadata are mapped into the inherited semantic representation,
from which song-level evidence is hierarchically aggregated into artist and
company profiles. Multidimensional company semantic fingerprints capture the
implemented organisational characteristics, while temporal analysis models
their stability, drift and evolution. Fingerprint evidence, temporal evidence
and qualified interpretation are subsequently integrated to infer
organisational semantic identities. Statistical, robustness, predictive,
reproducibility, traceability and validity analyses form a supporting
validation foundation across the framework rather than a single terminal
processing stage. Here, identity inference denotes corpus-dependent evidence synthesis; the predictive component is complementary and does not establish construct validity.
}
\label{fig:end_to_end_framework}

\end{figure}

Figure~\ref{fig:end_to_end_framework} brings together the framework’s main stages, from hierarchical aggregation and fingerprint construction to longitudinal comparison and qualified interpretation. Their evaluation is described in Section~\ref{sec:validation}.
\subsection{Inherited Semantic Representation}
\label{subsec:semantic_representation}
Each eligible song has an inherited multilingual representation
\[
\mathbf{z}_s \in \mathbb{R}^{d},
\]
where $d$ is its dimensionality. The fixed similarity graph and partition provide $K=15$ communities, denoted abstractly by
\[
g_s \in \{1,\ldots,K\}.
\]
These indices refer to the retained categories rather than their original identifiers. The figures preserve the labels C0--C13 and C15, without renumbering any results. Community membership provides a common basis for comparing portfolios, but does not itself define organisational identity.

\subsection{Hierarchical Aggregation}
\label{subsec:hierarchical_aggregation}
The hierarchy
\[
\text{lyrics}
\rightarrow
\text{songs}
\rightarrow
\text{artists}
\rightarrow
\text{companies}.
\]
retains release years throughout. For an artist's eligible songs $\mathcal{S}_a$, the profile
\[
\mathbf{p}_a
\]
contains the distributional and graph quantities used in aggregation and inference. A company is represented by
\[
\left\{
\mathbf{p}_a : a \in \mathcal{A}_c
\right\}.
\]
Retaining artist-level records makes it possible to examine variation within companies rather than relying solely on pooled portfolio totals. Period-specific song, artist and company-song subsets are
\[
\mathcal{S}_{a,t},
\qquad
\mathcal{A}_{c,t},
\qquad
\mathcal{S}_{c,t},
\]
respectively.

\subsection{Company Semantic Fingerprints}
\label{subsec:company_fingerprints}
For company $c$, the fingerprint is
\[
\mathbf{f}_c
=
\left[
f_{c}^{(1)},
f_{c}^{(2)},
\ldots,
f_{c}^{(M)}
\right]^{\top},
\]
where $M$ denotes the implemented dimensions. Composition describes community shares; diversity and concentration describe their breadth and balance; novelty describes representation-relative separation; brokerage and centrality describe graph relationships. Table~\ref{tab:fingerprint_dimensions} retains the canonical components and their interpretations.

The community-composition component is
\[
\boldsymbol{\pi}_c
=
\left[
\pi_{c1},
\ldots,
\pi_{cK}
\right]^{\top}
\]
where $\pi_{ck}$ is community $k$'s share of the eligible company portfolio and
\[
\pi_{ck} \geq 0,
\qquad
\sum_{k=1}^{K}\pi_{ck}=1.
\]
Because community composition cannot by itself describe graph position, variation among artists or change over time, the fingerprint is not reduced to a single identity score:
\[
\mathbf{f}_c
\neq
\text{a single organisational identity score}.
\]
\begin{table*}[t]
\centering
\caption{
Canonical components of the organisational semantic fingerprint.
Rather than representing an entertainment company by a single scalar score,
the proposed framework models each organisation through multiple complementary
semantic properties derived from the validated semantic-community structure.
Each component is operationalised using one or more implemented quantitative
metrics and collectively forms the multidimensional company semantic fingerprint.
}
\label{tab:fingerprint_dimensions}

\small
\renewcommand{\arraystretch}{1.25}

\begin{tabular}{p{2.2cm} p{4cm} p{6.8cm}}
\toprule

\textbf{Fingerprint Component}
&
\textbf{Implemented Metrics}
&
\textbf{Interpretation}
\\

\midrule

Community composition
&
Dominant semantic community,
dominant community proportion,
top-three community proportion
&
Describes how the company's lyrical catalogue is distributed across the validated semantic communities and identifies its principal semantic themes.
\\

\addlinespace

Semantic diversity
&
Normalised Shannon entropy,
effective number of communities,
Simpson diversity
&
Quantifies the breadth and balance of the company's semantic coverage. Higher values indicate a broader and more evenly distributed semantic portfolio.
\\

\addlinespace

Semantic concentration
&
Herfindahl--Hirschman Index (HHI)
&
Measures the extent to which lyrical output is concentrated within a limited number of semantic communities rather than being broadly distributed.
\\

\addlinespace

Graph integration
&
Mean participation coefficient,
mean neighbour entropy
&
Characterises how strongly the company's songs connect multiple semantic regions within the semantic similarity graph, indicating boundary-spanning behaviour and structural integration.
\\

\addlinespace

Semantic novelty
&
Mean centroid novelty,
mean local novelty
&
Measures the representation-relative novelty and distinctiveness of the company's semantic profile relative to the overall semantic landscape and its local neighbourhood.
\\

\addlinespace

Organisational scale
&
Number of eligible songs,
number of artists,
number of occupied semantic communities
&
Provides descriptive characteristics of the analysed catalogue and contextualises the remaining fingerprint dimensions without contributing directly to identity interpretation.
\\

\bottomrule
\end{tabular}

\end{table*}
\subsection{Temporal Modelling}
\label{subsec:temporal_modelling}
A company-period fingerprint
\[
\mathbf{f}_{c,t}
\]
forms part of the trajectory
\[
\mathcal{T}_c
=
\left\{
\mathbf{f}_{c,1},
\mathbf{f}_{c,2},
\ldots,
\mathbf{f}_{c,T_c}
\right\}
\]
across its $T_c$ observed periods. Consecutive change is
\[
D_{c,t}
=
d\!\left(
\mathbf{f}_{c,t},
\mathbf{f}_{c,t-1}
\right),
\]
for the implemented dissimilarity $d$. Stability is evaluated across this trajectory, recognising that fingerprint dimensions may evolve at different rates. The rolling-window, phase and stability-index definitions are provided alongside the results in Section~\ref{sec:temporal_evolution}; importantly, overlapping windows do not constitute independent observations.

\subsection{Qualified Identity Interpretation}
\label{subsec:identity_inference}
The identity interpretation
\[
\mathcal{I}_c,
\]
is organised as
\[
\mathcal{I}_c
=
\Phi
\left(
\mathbf{f}_c,
\mathcal{T}_c,
\mathcal{V}_c
\right),
\]
where $\mathcal{V}_c$ denotes validation and traceability evidence. $\Phi$ is evidence-synthesis notation, not a fitted estimator. Statements must trace to implemented metrics, remain comparative within the corpus, and retain temporal, uncertainty and sensitivity qualifications. Conceptually,
\[
\mathcal{I}_c
=
\left\{
\begin{tabular}{l}
\text{dominant semantic characteristics},\\
\text{distinguishing features},\\
\text{temporal qualification},\\
\text{evidential qualification}
\end{tabular}
\right\}.
\]
These requirements make the interpretation more accountable to the measured evidence, although they cannot establish broader construct validity.

For the integrated display, the retained implementation standardises each input across the four companies using population-standard-deviation z-scores, reverses negatively oriented inputs, and averages equally within each dimension. Exploration combines mean centroid novelty, mean local novelty and relative organisational distinctiveness. Integration combines mean bridge affinity, participation coefficient and maximum centroid cosine. Breadth combines normalised semantic entropy, reversed HHI and community coverage. Adaptation combines mean JS divergence and reversed mean cosine similarity. These relative composites are distinct from any single canonical metric.

Displayed signature strength averages absolute robust z-scores across the 21 retained signature components; robust scores multiply the deviation from the across-company median by 0.6744897501960817 and divide by the median absolute deviation, with ordinary z-scores as the zero-deviation fallback. Leading communities are the three highest-ranked signed contribution scores: company community share multiplied by the base-two logarithm of its ratio to the pooled community share. They are not necessarily the largest communities or statistically significant enrichments. Signature components include temporal and recovery diagnostics, so strength is an evidence summary rather than independent corroboration. The resulting labels describe relative configurations, not motivations or corporate values.

\section{Evaluation of the Computational Framework}
\label{sec:validation}
\subsection{Strategy}
\label{subsec:validation_strategy}
\begin{figure*}[!t]
    \centering
    \includegraphics[width=\textwidth]{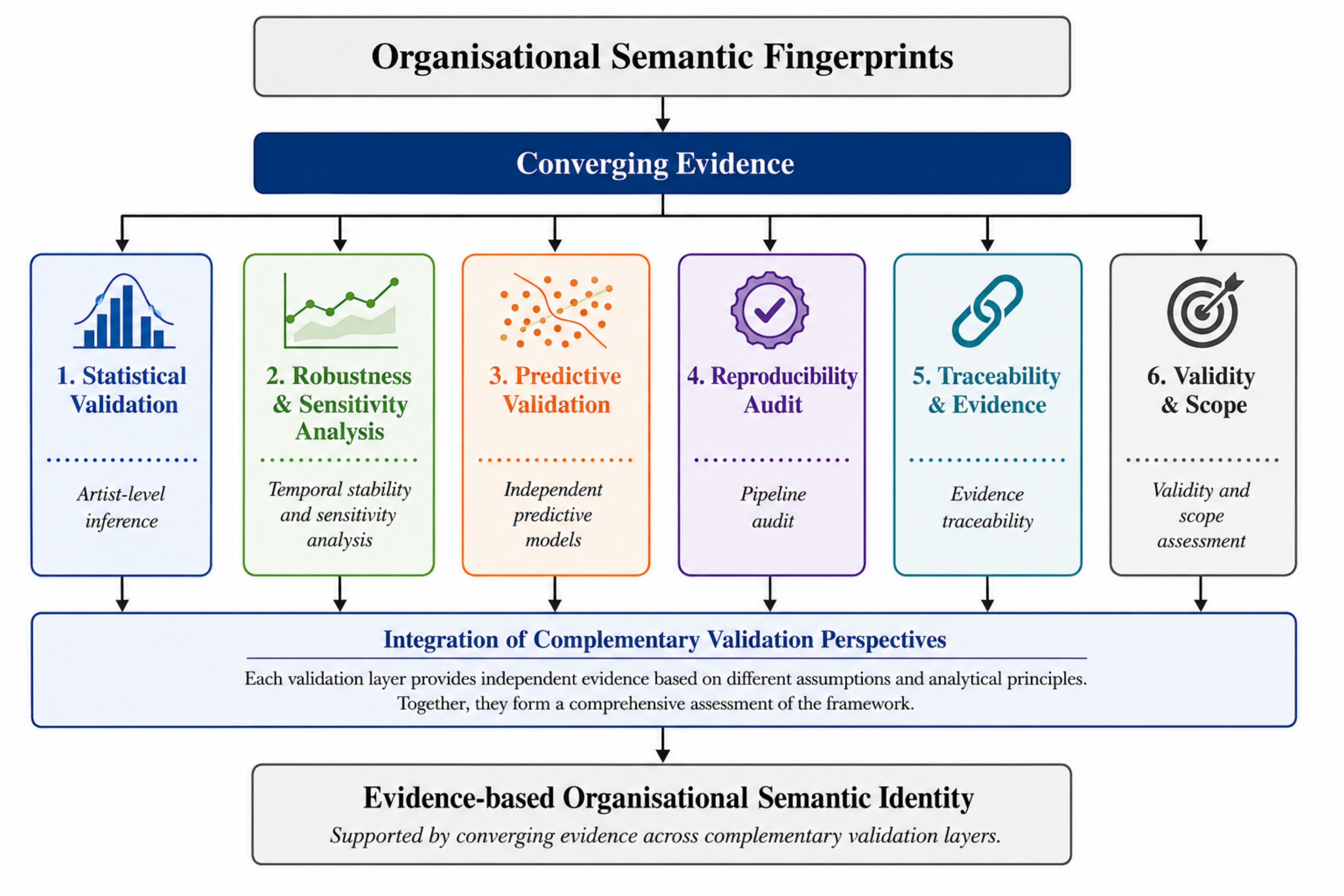}
    \caption{\textbf{Multi-layer validation architecture supporting the proposed computational framework.} The organisational semantic fingerprints produced by the proposed framework are evaluated through six complementary validation perspectives: (1) statistical validation, (2) robustness and sensitivity analysis, (3) predictive validation, (4) reproducibility auditing, (5) evidence traceability, and (6) validity and scope assessment. The layers assess distinct properties and limitations of the measured fingerprints. The figure's predictive-validation label denotes a complementary diagnostic, not primary validation of semantic identity; convergence within the same corpus does not establish external construct validity.}
    \label{fig:validation_framework}
\end{figure*}

We assess the corpus-dependent interpretations from six complementary perspectives: statistical comparison, robustness, prediction, reproducibility, traceability and scope. Because all six draw on the same corpus and semantic foundation, agreement among them should not be mistaken for independent external validation of the construct. Figure~\ref{fig:validation_framework} summarises their respective roles.
\subsection{Statistical Comparison}
\label{subsec:statistical_validation}
Let
\[
x_{ai}^{(m)}
\]
denote component $m$ for artist $a$ assigned to company $i$. Artists, rather than pooled company summaries, are the inferential units. Non-parametric tests accommodate bounded distributions and graph-derived measures without assuming Gaussian observations. Kruskal--Wallis tests assess
\[
H_0: F_1 = F_2 = \cdots = F_C,
\]
where $F_i$ is the artist-level distribution and $C$ the number of companies. Mann--Whitney pairwise comparisons follow detected overall differences; Holm adjustment controls multiple comparisons. Rank-biserial correlation
\[
r_{\mathrm{rb}} \in [-1,1],
\]
reports pairwise magnitude and direction alongside adjusted significance. A statistically significant result or descriptive contrast does not, on its own, establish a substantive organisational distinction. Conversely, a non-significant result should not be interpreted as evidence of equivalence.

\subsection{Robustness and Sensitivity}
\label{subsec:robustness_validation}
Six retained analyses compare similarity metrics, bootstrap profile reconstruction against random ordering, leave-one-artist-out exclusions, balanced catalogue sampling, alternative temporal windows and alternative community partitions. We assess continuous geometric variation separately from exact rank retention, since even modest numerical changes can affect the ordering of portfolios. Temporal sensitivity compares company-period fingerprints through
\[
D_{c,t}=d\!\left(\mathbf{f}_{c,t},\mathbf{f}_{c,t-1}\right),
\]
under annual and alternative multi-year windows. Section~\ref{sec:framework_validation} reports the quantitative anchors.

\subsection{Complementary Prediction, Recovery and Persistence}
\label{subsec:predictive_validation}
Affiliation prediction uses repeated stratified song-level and artist-grouped validation, with five folds and ten repeats; the latter keeps each artist's songs together. Retained feature sets comprise community membership, seven graph-derived quantities, and their combination. Logistic regression and random forest models are compared with dummy baselines; the reported headline scores use logistic regression with combined features. Numeric median imputation and standardisation, categorical imputation and one-hot encoding, and classifier fitting occur within training folds. The inherited embeddings, graph, communities and derived input features remain fixed rather than being reconstructed in each fold. Thus artist separation evaluates classification on a shared corpus representation, not a fully independent representation-learning pipeline. Balanced accuracy and macro-F1 accompany the four-class baseline of 0.25. The reported 95\% intervals are empirical fold-score percentiles, not confidence intervals for an independent-sample mean.

Catalogue recovery samples without replacement at each fraction, with 250 repetitions, and compares sample fingerprints against full-company references that include the sampled songs. Graph standardisation uses the complete eligible population. Assignment minimises an equally weighted combination of graph root-mean-square distance and square-root community JS divergence. Recovery measures reconstruction of known portfolios, not held-out company identification.

Temporal assessment compares persistence, linear-trend and moving-average forecasts using expanding history over rolling windows. Retained settings require at least twelve songs and two distinct years per window; the moving average uses three historical fingerprints. All models use common targets beginning with the third available fingerprint, and graph standardisation uses available history only. Successive three-, five- and seven-year windows overlap, so targets share observations with preceding windows. Community cosine similarity therefore measures persistence in overlapping representations, not independent future-catalogue forecasting. These diagnostics provide supporting evidence for the interpretation of semantic identity; they do not define it.

\subsection{Reproducibility, Traceability and Scope}

The reported reproducibility audit checks retained aggregation,
statistical, temporal, robustness, predictive and synthesis stages,
including execution success, controlled random states and intermediate
outputs. The reported outcome belongs to the existing analysis and
should not be understood as a new audit undertaken for this revision.

Evidence traceability follows the analytical hierarchy from lyrics
and songs through artists and companies to semantic fingerprints and
their qualified organisational interpretation. Each interpretation
is linked to the corresponding fingerprint, comparative and temporal
evidence, allowing the analytical chain to be inspected without
implying a deterministic organisational explanation.

Statistical validity concerns observational units, corrections and
effect sizes; computational validity concerns implementation and
reproducibility; data validity concerns eligibility, static affiliation
assignment, coverage and representation dependence. Construct
validity concerns the relationship between lyrical patterns and
broader organisational concepts. The study remains descriptive and
is limited to the observed corpus and four company portfolios.
It cannot identify managerial intent or causal effects of company
affiliation, and application to another setting would require fresh
assessment of both the domain-specific representation and the
proposed construct.
\section{Results}
\label{sec:results}

\subsection{Organisation-level Semantic Fingerprints}
\label{sec:semantic_fingerprints}

\begin{figure*}[t]
    \centering
    \includegraphics[width=\textwidth]{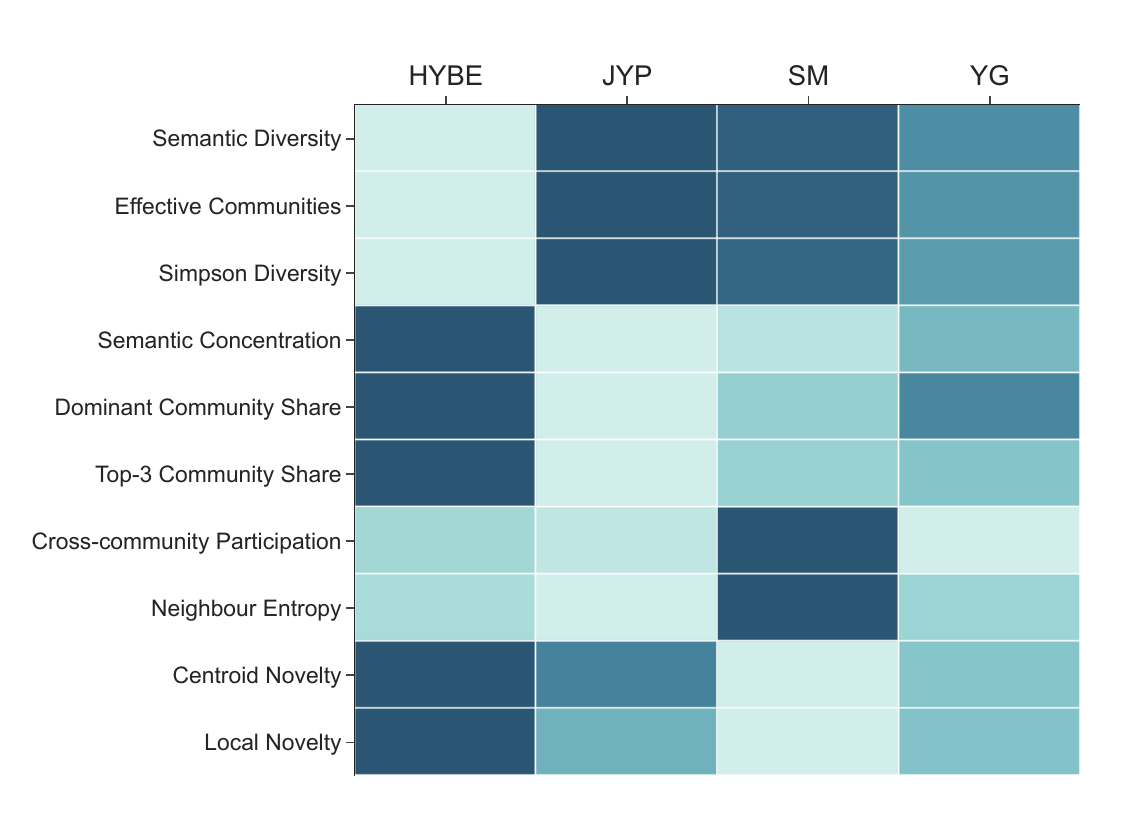}
    \caption{\textbf{Organisation-level semantic fingerprints of the four entertainment companies.} The heatmap compares HYBE, JYP, SM and YG across the ten canonical semantic dimensions. Values are scaled independently within each dimension to emphasise relative company-level differences; consequently, colour intensity should be interpreted within rows rather than compared across different metrics. The figure shows descriptive profile differences spanning diversity, concentration, connectivity and novelty; artist-level statistical support is assessed separately and is selective.}
    \label{fig:company_semantic_fingerprint_dashboard}
\end{figure*}

\begin{figure*}[t]
    \centering
    \includegraphics[width=\textwidth]{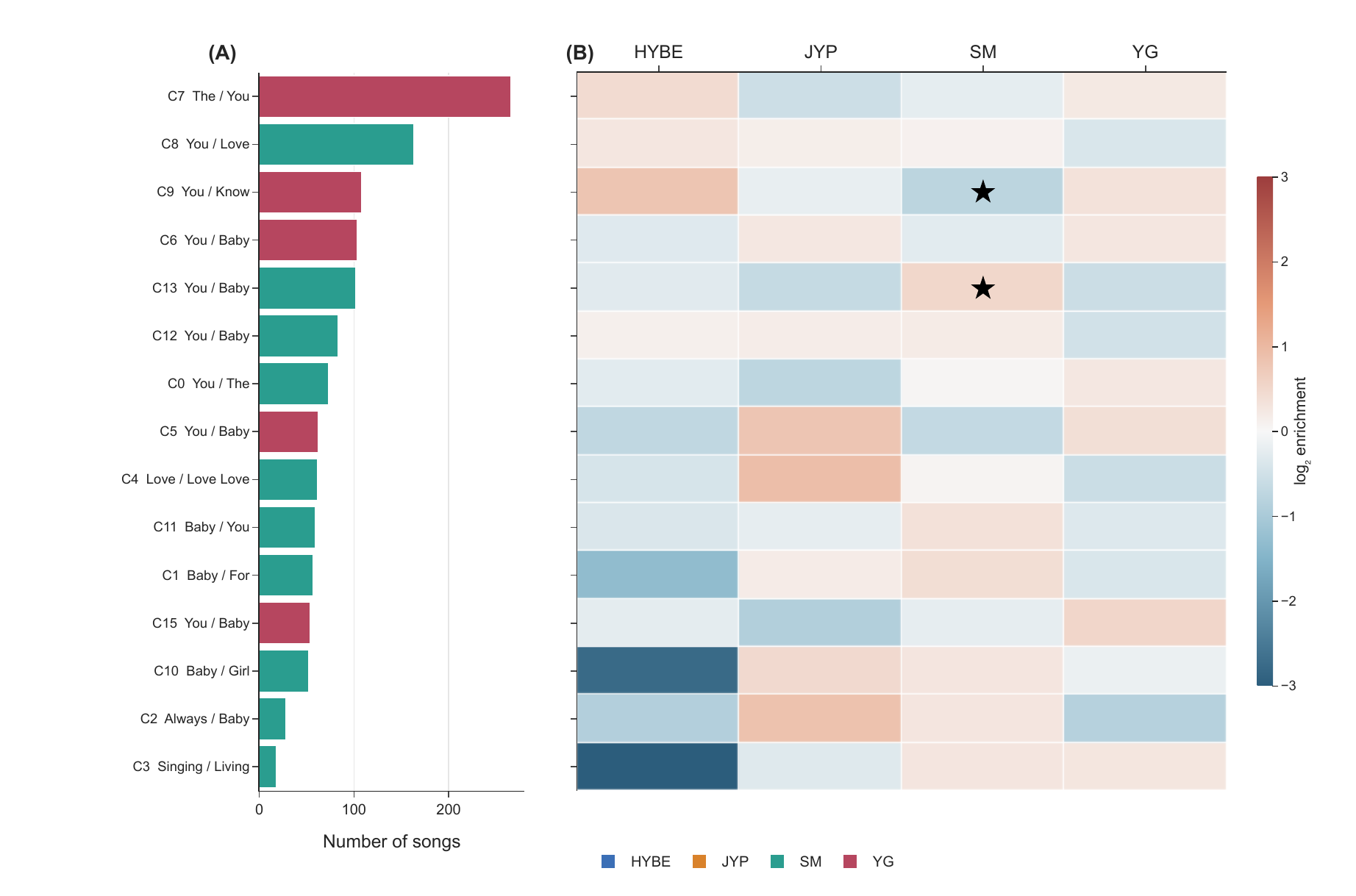}
    \caption{\textbf{Semantic community prevalence and company enrichment.} (A) Number of songs in each semantic community, ordered from largest to smallest and coloured by the company contributing the largest share of songs. (B) Company-by-community enrichment expressed as log$_2$ enrichment ratios. Stars denote associations that remain significant after Holm correction. Community labels show the first two evidence terms; complete three-term descriptors are reported in Table~\ref{tab:semantic_community_summary}.}    \label{fig:semantic_community_landscape}
\end{figure*}

The organisation-level fingerprints are compared in Figure~\ref{fig:company_semantic_fingerprint_dashboard}, with their underlying semantic communities shown in Figure~\ref{fig:semantic_community_landscape}. Tables~\ref{tab:company_fingerprint_summary} and~\ref{tab:semantic_community_summary} provide the numerical profiles and community reference information.

The fingerprint comparison reveals distinct descriptive tendencies. JYP has the highest Semantic Diversity, Effective Communities and Simpson Diversity, indicating the broadest distribution across the inherited communities, whereas HYBE leads in Semantic Concentration, Dominant Community Share and Top-3 Community Share, reflecting a more concentrated portfolio. SM has the highest Cross-community Participation and Neighbour Entropy, consistent with greater connectivity across communities. HYBE also records the highest Centroid Novelty and Local Novelty within the learned embedding space, while YG generally occupies intermediate positions. These portfolio-level contrasts are descriptive; their artist-level statistical support is examined separately.

\begin{table}[t]
\centering
\caption{Organisation-level semantic fingerprint summary of the four
entertainment companies. Values correspond to the canonical semantic
dimensions visualised in Figure~\ref{fig:company_semantic_fingerprint_dashboard}.
Boldface indicates the highest observed value for each semantic dimension.}
\label{tab:company_fingerprint_summary}

\begin{threeparttable}

\small
\setlength{\tabcolsep}{6pt}
\renewcommand{\arraystretch}{1.20}

\sisetup{
    detect-weight = true,
    detect-inline-weight = math
}

\begin{tabular}{
    ll
    S[table-format=2.3]
    S[table-format=2.3]
    S[table-format=2.3]
    S[table-format=2.3]
}

\toprule
\textbf{Category} &
\textbf{Semantic Dimension} &
{\textbf{HYBE}} &
{\textbf{JYP}} &
{\textbf{SM}} &
{\textbf{YG}} \\
\midrule

\multirow{3}{*}{\textbf{Diversity}}
& Semantic Diversity
& 0.821
& {\bfseries 0.942}
& 0.935
& 0.903 \\

& Effective Communities
& 9.234
& {\bfseries 12.820}
& 12.584
& 11.534 \\

& Simpson Diversity
& 0.855
& {\bfseries 0.912}
& 0.907
& 0.889 \\

\midrule

\multirow{3}{*}{\textbf{Concentration}}
& Semantic Concentration
& {\bfseries 0.145}
& 0.088
& 0.093
& 0.111 \\

& Dominant Community Share (\%)
& {\bfseries 28.2}
& 14.0
& 17.6
& 24.2 \\

& Top-3 Community Share (\%)
& {\bfseries 58.2}
& 37.6
& 42.6
& 44.6 \\

\midrule

\multirow{2}{*}{\textbf{Connectivity}}
& Cross-community Participation
& 0.550
& 0.544
& {\bfseries 0.589}
& 0.541 \\

& Neighbour Entropy
& 0.402
& 0.394
& {\bfseries 0.438}
& 0.404 \\

\midrule

\multirow{2}{*}{\textbf{Novelty}}
& Centroid Novelty
& {\bfseries 0.044}
& 0.040
& 0.028
& 0.033 \\

& Local Novelty
& {\bfseries 0.056}
& 0.044
& 0.034
& 0.042 \\

\bottomrule
\end{tabular}

\begin{tablenotes}[flushleft]
\footnotesize
\item \textit{Notes.}
Semantic Diversity denotes normalised Shannon entropy, while Semantic
Concentration denotes the Herfindahl--Hirschman concentration index.
Higher values indicate greater diversity, cross-community participation,
neighbour entropy or novelty, depending on the corresponding dimension.
Higher Semantic Concentration, Dominant Community Share and Top-3 Community
Share indicate a more concentrated semantic portfolio rather than superior
performance. Shares are reported as percentages. Company order is fixed as
HYBE, JYP, SM and YG throughout the manuscript.
\end{tablenotes}

\end{threeparttable}
\end{table}

Figure~\ref{fig:semantic_community_landscape} examines the semantic structures underlying the organisation-level fingerprints. Panel~(A) summarises the distribution of songs across the inherited semantic communities, ordered by community size and coloured according to the company contributing the largest proportion of songs. Although the largest contributor varies between communities, all major communities contain songs from multiple companies, indicating that the organisational fingerprints are constructed from a shared semantic landscape rather than from company-exclusive semantic topics.

Panel~(B) compares company-level enrichment across the semantic communities. Most enrichment values remain close to zero, indicating relatively balanced representation across companies. Following Holm correction, only two statistically significant associations are identified: Community~C13 is significantly over-represented for SM, whereas Community~C9 is significantly under-represented for SM. No other company--community combinations remain significant after multiple-comparison correction, suggesting that the observed organisation-level differences arise primarily through gradual shifts in the relative weighting of shared semantic communities rather than through widespread company-specific semantic specialisation.

\begin{table}[t]
\centering
\caption{Reference summary of the semantic communities underlying the
organisation-level semantic fingerprints. Communities are ordered by size,
consistent with Figure~\ref{fig:semantic_community_landscape}. Representative
vocabulary comprises the three highest-ranked evidence terms for each
community.}
\label{tab:semantic_community_summary}

\begin{threeparttable}

\small
\setlength{\tabcolsep}{6pt}
\renewcommand{\arraystretch}{1.18}

\begin{tabular}{clcll}

\toprule
\textbf{Community} &
\textbf{Representative Vocabulary} &
\textbf{Songs} &
\textbf{Largest Contributor} &
\textbf{Notable Association} \\
\midrule

C7  & The / You / Like
    & 266 & YG (37.6\%) & None \\

C8  & You / Love / Baby
    & 163 & SM (44.2\%) & None \\

C9  & You / Know / Don
    & 108 & YG (40.7\%) & SM under-represented \\

C6  & You / Baby / For
    & 103 & YG (38.8\%) & None \\

C13 & You / Baby / The
    & 102 & SM (58.8\%) & SM over-represented \\

C12 & You / Baby / Let
    & 83 & SM (47.0\%) & None \\

C0  & You / The / Baby
    & 73 & SM (42.5\%) & None \\

C5  & You / Baby / Don
    & 62 & YG (41.9\%) & None \\

C4  & Love / Love Love / You
    & 61 & SM (42.6\%) & None \\

C11 & Baby / You / Take
    & 59 & SM (52.5\%) & None \\

C1  & Baby / For / Sorry
    & 57 & SM (54.4\%) & None \\

C15 & You / Baby / Home
    & 54 & YG (46.3\%) & None \\

C10 & Baby / Girl / Boy
    & 52 & SM (50.0\%) & None \\

C2  & Always / Baby / For
    & 28 & SM (50.0\%) & None \\

C3  & Singing / Living / Love
    & 18 & SM (50.0\%) & None \\

\bottomrule
\end{tabular}

\begin{tablenotes}[flushleft]
\footnotesize
\item \textit{Notes.}
Communities are ordered by decreasing number of songs to match
Figure~\ref{fig:semantic_community_landscape}. Representative vocabulary
comprises the three highest-ranked lexical evidence terms derived from the
community semantic dictionary; these terms are reproducible descriptors rather
than manually assigned topic labels. Largest Contributor identifies the company
contributing the greatest proportion of songs within each community. Notable
Association reports statistically significant company--community enrichment
after Holm correction at $p<0.05$. ``None'' indicates that no significant
over- or under-representation was detected.
\end{tablenotes}

\end{threeparttable}
\end{table}

\subsection{Statistical Differentiation of Artist-level Semantic Profiles}
\label{sec:artist_level_statistics}

\begin{figure*}[t]
    \centering
    \includegraphics[width=\textwidth]{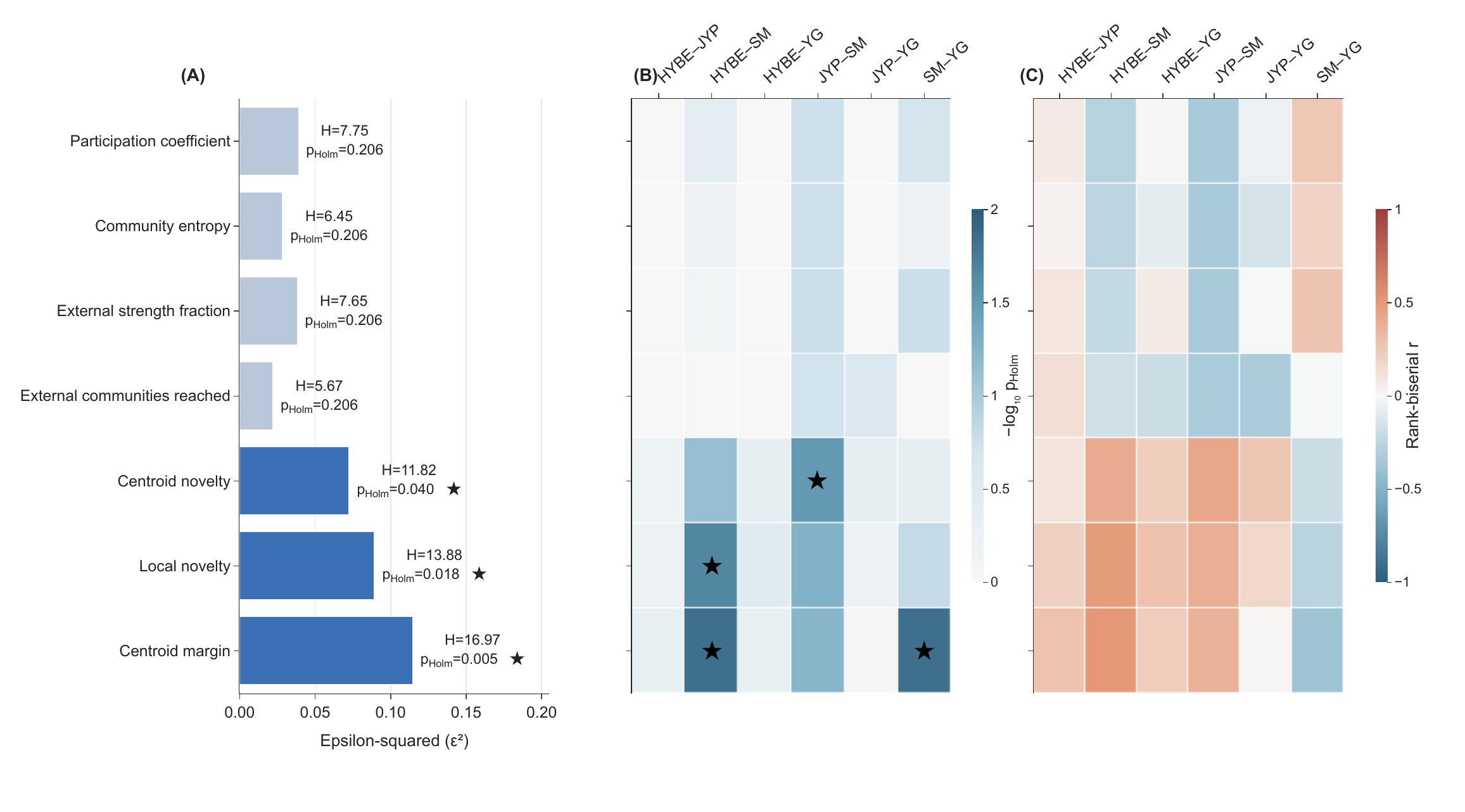}
    \caption{
    \textbf{Statistical differentiation of artist-level semantic metrics across the Big Four entertainment companies.}
    (A) Global Kruskal--Wallis tests summarising differences across all companies. Horizontal bars represent the corresponding epsilon-squared ($\varepsilon^{2}$) effect sizes, with annotations reporting the Kruskal--Wallis statistic ($H$) and Holm-adjusted $p$-value. Metrics remaining statistically significant after Holm correction are indicated by a star ($\star$). (B) Holm-corrected pairwise Mann--Whitney comparisons between companies. Cell colour represents $-\log_{10}(p_{\mathrm{Holm}})$, while stars denote statistically significant comparisons ($\alpha=0.05$). (C) Rank-biserial correlation coefficients for the corresponding pairwise comparisons, illustrating both the magnitude and direction of the observed effects. Positive values indicate larger metric values for the first company in each comparison, whereas negative values indicate larger values for the second company.
    }
    \label{fig:statistical_differentiation}
\end{figure*}

Artist-level Kruskal--Wallis tests, Holm-corrected Mann--Whitney comparisons and rank-biserial effect sizes assess whether descriptive portfolio differences are supported among contributors (Figure~\ref{fig:statistical_differentiation}; Table~\ref{tab:statistical_summary}).

Figure~\ref{fig:statistical_differentiation}(A) reveals a clear distinction between descriptive and inferential evidence. Four metrics describing semantic diversity and cross-community connectivity (Participation Coefficient, Community Entropy, External Strength Fraction, and External Communities Reached) did not remain statistically significant after Holm correction despite exhibiting observable descriptive differences across companies. This suggests that variation in these properties is modest relative to the within-company variability observed among individual artists.

In contrast, all three novelty-oriented metrics remained statistically significant after correction for multiple comparisons. Centroid Novelty ($H=11.82$, $\varepsilon^{2}=0.072$, $p_{\mathrm{Holm}}=0.040$), Local Novelty ($H=13.88$, $\varepsilon^{2}=0.089$, $p_{\mathrm{Holm}}=0.018$), and Centroid Margin ($H=16.97$, $\varepsilon^{2}=0.114$, $p_{\mathrm{Holm}}=0.005$) all demonstrated moderate effect sizes. These results indicate that the primary semantic distinctions between companies are associated with the novelty and separation of artist semantic profiles rather than with broader differences in semantic diversity or network connectivity.

The pairwise comparisons presented in Figure~\ref{fig:statistical_differentiation}(B--C) further localise these differences. Only four company pairs remained statistically significant following Holm correction. JYP and SM differed significantly in Centroid Novelty, indicating differences in the global distinctiveness of artist semantic profiles. HYBE and SM differed in both Local Novelty and Centroid Margin, suggesting systematic differences in local semantic uniqueness and neighbourhood separation. Finally, Centroid Margin also differed significantly between SM and YG, demonstrating that these organisations exhibit different degrees of semantic separation although the tested diversity and connectivity differences did not remain significant after correction.

Taken together, these results demonstrate that statistically robust organisational differences are concentrated in semantic novelty rather than semantic diversity. The strongest inferential support for the descriptive fingerprints in Section~\ref{sec:semantic_fingerprints} therefore concerns the distinctiveness and separation of artist semantic profiles. The non-significant diversity and connectivity tests do not establish equivalence; they indicate that systematic differences were not detected under the specified tests and correction. Descriptive contrasts and within-company heterogeneity remain relevant. A summary of these inferential findings is provided in Table~\ref{tab:statistical_summary}.

\begin{table*}[t]
\centering
\caption{
Summary of inferential statistical comparisons across artist-level semantic profile metrics. Statistically significant differences after Holm correction are concentrated in novelty-related measures, whereas the tested diversity and cross-community connectivity differences do not remain significant after correction.
}
\label{tab:statistical_summary}

\small

\begin{tabular}{p{2.5cm}p{1.2cm}p{2.6cm}p{1.2cm}p{4.0cm}}
\toprule
\textbf{Metric} &
\textbf{Global} &
\textbf{Significant Pair(s)} &
\textbf{Largest $|r|$} &
\textbf{Interpretation} \\
\midrule

Participation Coefficient &
No &
-- &
Small &
No statistically significant company difference in cross-community participation was detected after correction. \\

Community Entropy &
No &
-- &
Small &
Semantic diversity within artist neighbourhoods does not differ significantly. \\

External Strength Fraction &
No &
-- &
Small &
No statistically significant company difference in external semantic connectivity was detected after correction. \\

External Communities Reached &
No &
-- &
Small &
No statistically significant company difference in external communities reached was detected after correction. \\

Centroid Novelty &
Yes &
JYP--SM &
0.43 &
Companies differ in the global distinctiveness of artist semantic profiles. \\

Local Novelty &
Yes &
HYBE--SM &
0.49 &
Local semantic uniqueness differs significantly between companies. \\

Centroid Margin &
Yes &
HYBE--SM; SM--YG &
0.51 &
The degree of semantic separation from neighbouring identities varies significantly across companies. \\

\bottomrule
\end{tabular}
\end{table*}

\subsection{Temporal Evolution of Organisational Semantic Identity}
\label{sec:temporal_evolution}

To investigate how organisational semantic identity evolves over time, we constructed temporal company representations using overlapping five-year rolling windows advanced in one-year increments. For each company, all songs released within a given five-year interval were aggregated to estimate the organisation's temporal semantic profile during that period, represented by the probability distribution over the learned semantic communities introduced in Section~\ref{subsec:semantic_representation}. Consequently, successive temporal observations correspond to windows such as \textit{1997--2001}, \textit{1998--2002}, and \textit{1999--2003}, rather than individual calendar years. This representation provides temporally aggregated semantic profiles while preserving the ability to detect gradual organisational evolution across adjacent periods.

Temporal semantic evolution was quantified exclusively through \emph{consecutive} rolling-window transitions. For each company, Jensen--Shannon (JS) divergence was computed between neighbouring five-year semantic profiles (e.g., \textit{1997--2001} $\rightarrow$ \textit{1998--2002}), thereby measuring incremental semantic change between highly overlapping organisational representations. Transitions separated by missing temporal windows were excluded from the consecutive analysis to ensure consistent temporal continuity. The resulting sequence of consecutive JS divergences forms the basis for trajectory analysis, temporal trend estimation, change-point detection, and the composite stability measures presented below. For phase-level comparison, each company's chronologically ordered sequence of consecutive semantic transitions was partitioned into three approximately equal-sized segments, denoted as the early, middle, and late phases. These phases therefore represent relative stages of each company's observed semantic trajectory rather than common calendar periods shared across organisations.

\begin{figure*}[t]
    \centering
    \includegraphics[width=\textwidth]{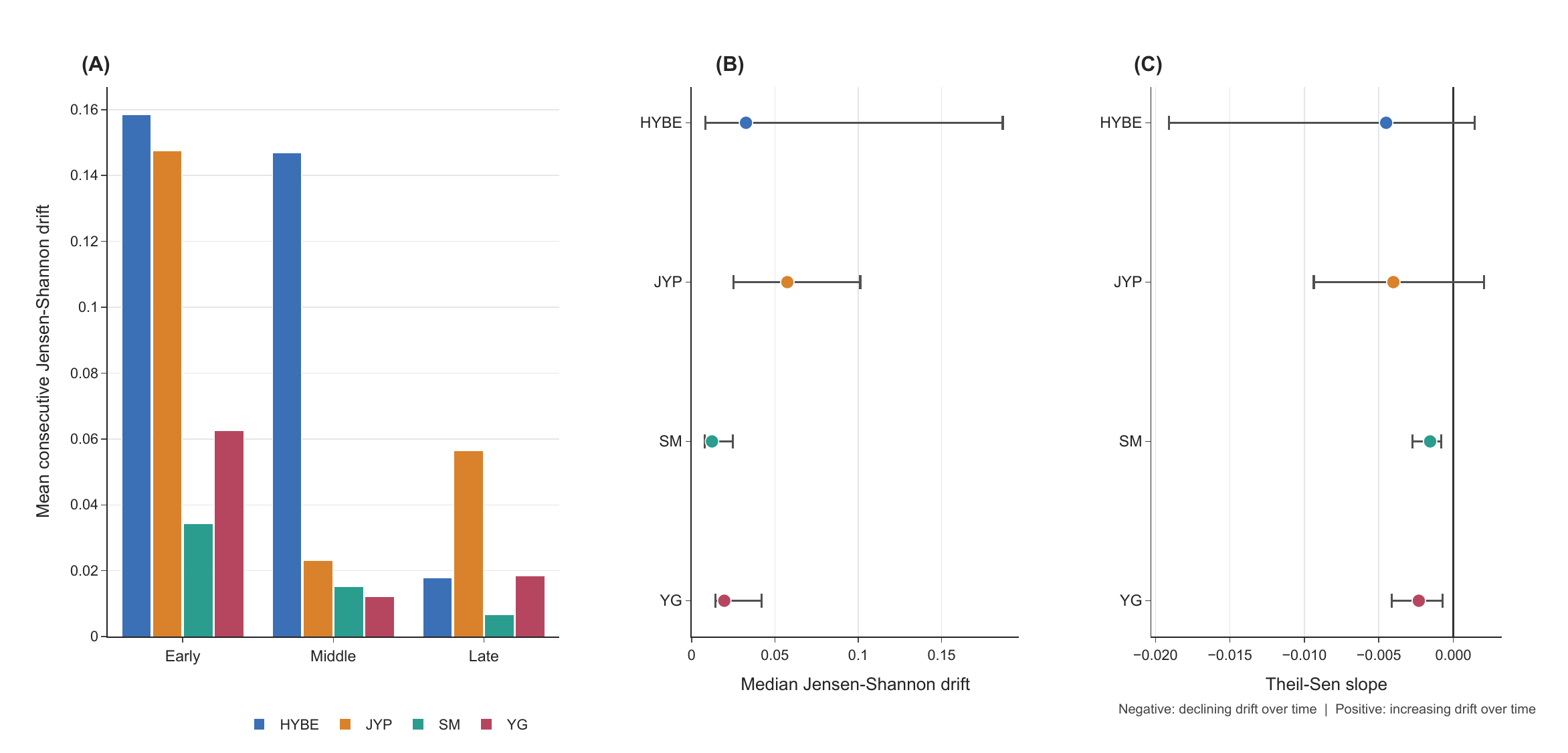}
    \caption{Temporal evolution of organisational semantic identity derived from overlapping five-year rolling windows. (A) Mean consecutive Jensen--Shannon semantic drift across the early, middle, and late phases of each company's temporal history. (B) Typical magnitude of consecutive semantic drift summarised by the median and interquartile range. (C) Long-term temporal trend estimated using the Theil--Sen slope and its confidence interval, where negative values indicate progressively smaller semantic changes over time.}
    \label{fig:temporal_evolution}
\end{figure*}

Figure~\ref{fig:temporal_evolution}(A) compares mean consecutive drift across relative phases. HYBE declines from early ($0.159$) and middle ($0.147$) to late drift ($0.018$). JYP follows a non-monotonic trajectory ($0.148$, $0.023$, $0.056$). SM remains low ($0.034 \rightarrow 0.015 \rightarrow 0.007$); YG shows moderate early movement followed by lower drift ($0.063 \rightarrow 0.012 \rightarrow 0.019$).

Typical displacement differs from maximum instability (Figure~\ref{fig:temporal_evolution}B). JYP has the largest median JS divergence (0.057). HYBE has a lower median (0.033) but the widest interquartile range. SM and YG have smaller medians (0.012 and 0.020, respectively).

Theil--Sen slopes are negative for all companies (Figure~\ref{fig:temporal_evolution}C). SM ($-0.0016$) and YG ($-0.0023$) have confidence intervals entirely below zero. HYBE ($-0.0045$) and JYP ($-0.0040$) have intervals crossing zero, providing weaker evidence of sustained decline in drift. Overlapping observations constrain these temporal interpretations.

\begin{figure*}[t]
    \centering
    \includegraphics[width=\textwidth]{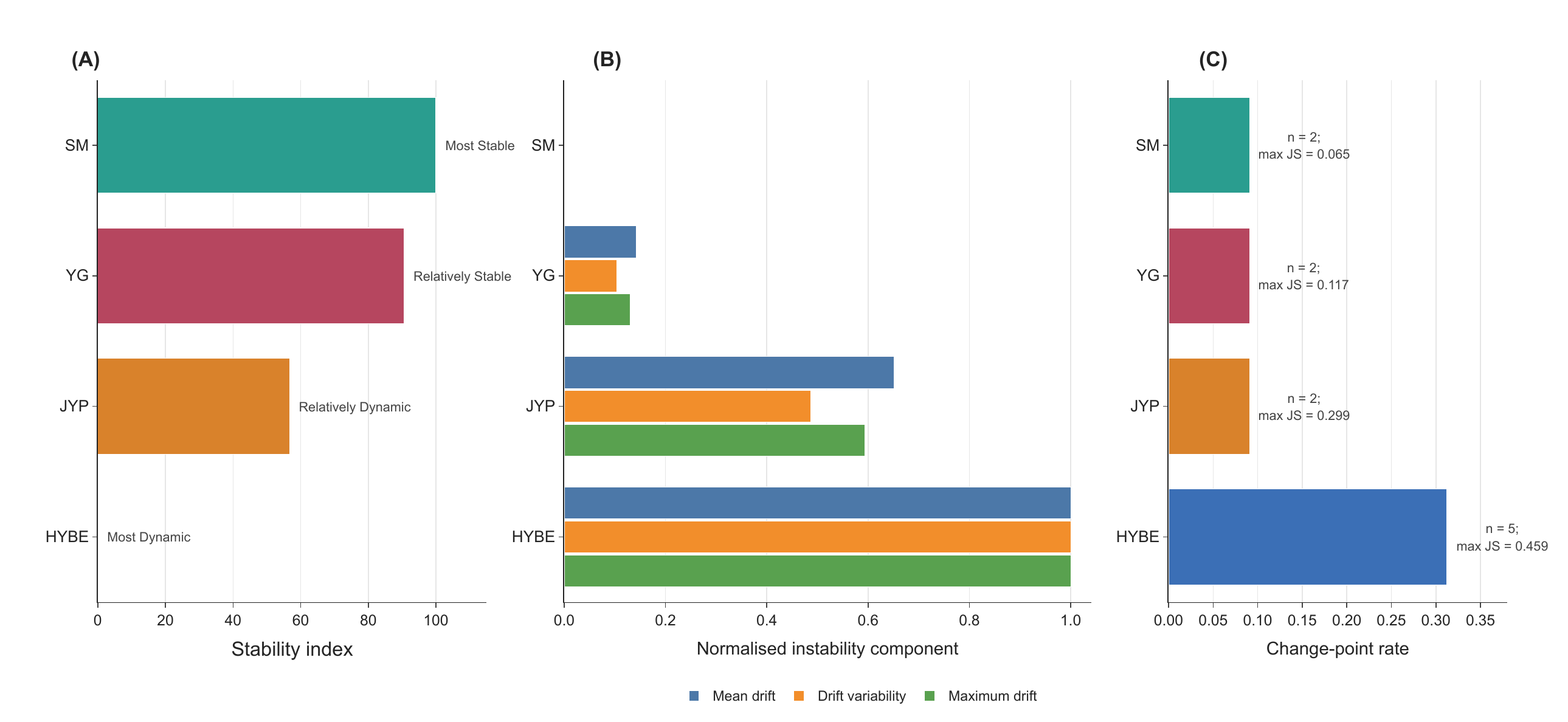}
    \caption{Temporal stability of organisational semantic identity. (A) Composite stability index derived from normalised measures of mean semantic drift, drift variability, maximum semantic drift, and change-point frequency. Higher values indicate greater long-term semantic stability. (B) Min--max-normalised values of the principal drift-based instability components: mean drift, drift variability, and maximum drift. (C) Frequency of semantic change points relative to the number of consecutive temporal transitions. Labels indicate the absolute number of detected change points and the maximum Jensen--Shannon divergence associated with those events.}
    \label{fig:temporal_stability}
\end{figure*}

Composite stability is highest for SM (100.0), followed by YG (90.6), JYP (56.7) and HYBE (0.0) (Figure~\ref{fig:temporal_stability}A). These are relative four-company indices, not absolute stability thresholds.

HYBE has the greatest mean drift, variability and maximum drift; JYP has elevated average and maximum drift with lower variability. SM is lowest across these components and YG is similarly low (Figure~\ref{fig:temporal_stability}B). The aggregate contrast does not imply continuous change: HYBE's late-period drift declines.

HYBE has five detected change points, approximately 31\% of consecutive comparisons, and maximum JS = 0.459 (Figure~\ref{fig:temporal_stability}C). JYP, SM and YG each have two, approximately 9\% of transitions. Maximum event magnitudes are greater for JYP (JS = 0.299) than YG (0.117) or SM (0.065). Thus frequency and magnitude need not rank portfolios identically.

\subsection{Integrated Organisational Semantic Identity Profiles}
\label{sec:integrated_identity_profiles}

\begin{figure*}[t]
    \centering
    \includegraphics[width=\textwidth]{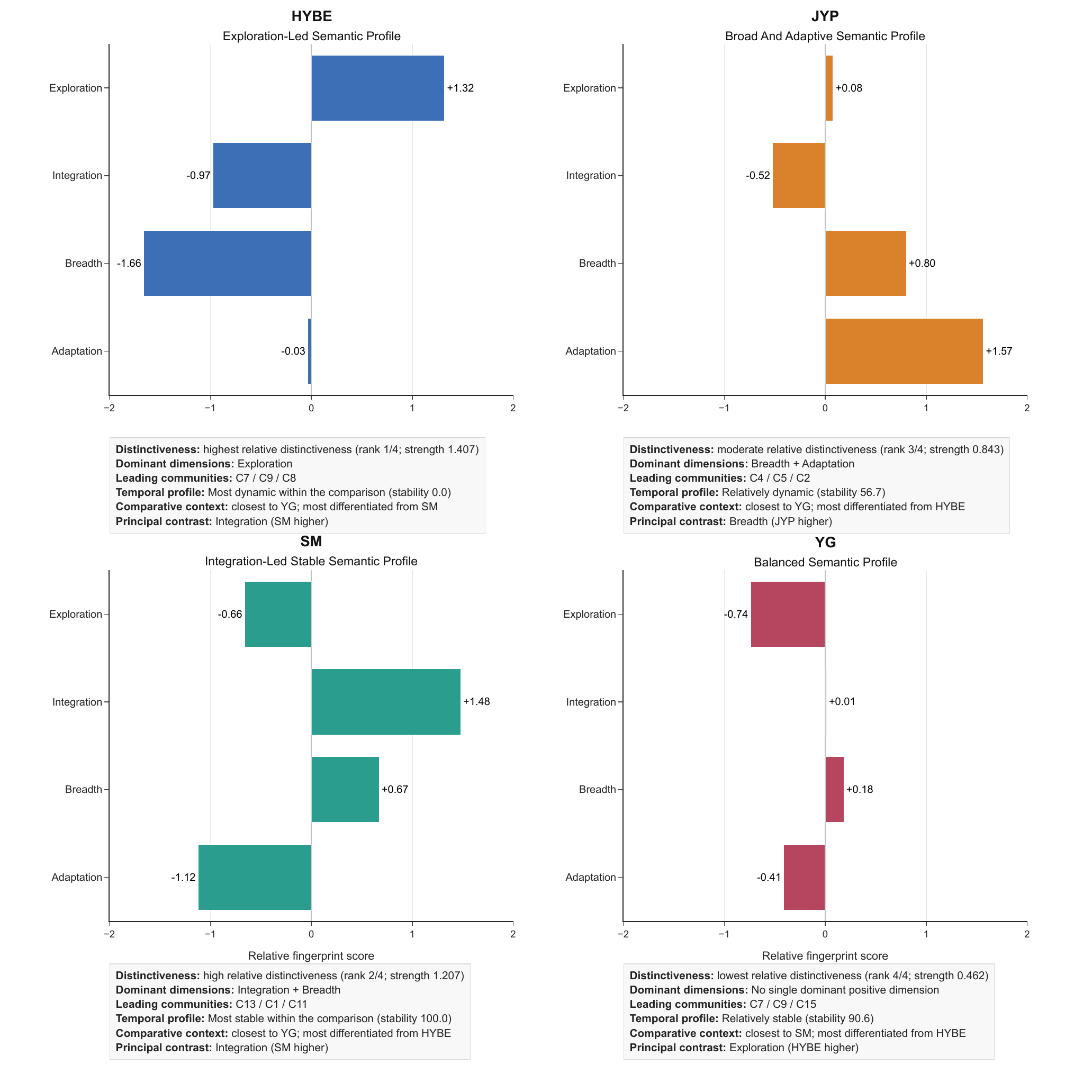}
    \caption{
    \textbf{Integrated organisational semantic identity profiles of the four entertainment companies.}
    Each panel summarises the complete evidence supporting the inferred organisation-level semantic identity. The horizontal bars represent the relative structural fingerprint across the four principal semantic dimensions (exploration, integration, breadth and adaptation). The accompanying evidence panel reports the organisation's relative semantic distinctiveness, dominant fingerprint dimensions, leading semantic communities, temporal qualification derived from the rolling five-year analysis, and comparative semantic relationships within the four-company landscape. Together, these complementary evidence sources provide an interpretable, traceable synthesis of organisation-level semantic identity inferred from the analysed lyrical corpus.
    }
    \label{fig:integrated_identity_profiles}
\end{figure*}

Figure~\ref{fig:integrated_identity_profiles} combines structural, community, comparative and temporal evidence using the mappings in Section~\ref{subsec:identity_inference}. Exploration, integration, breadth and adaptation are relative computational properties, not inferred motivations. Their descriptive labels do not imply significant artist-level separation.

HYBE has the highest integrated distinctiveness and exploration, with comparatively low breadth and integration. Its leading communities are C7, C9 and C8. It has the lowest relative temporal stability and greatest maximum displacement; declining late-period drift qualifies its exploration-led, variable profile.

JYP's dominant integrated dimensions are breadth and adaptation, with leading communities C4, C5 and C2. Its temporal variability exceeds SM and YG while remaining below HYBE on the composite stability comparison. Its profile combines broad coverage with measured temporal adaptation.

SM's descriptive profile emphasises integration and breadth, with leading communities C13, C1 and C11 and the strongest temporal stability. Its integration-led description reflects comparative continuity rather than an inferred organisational strategy.

YG has the lowest integrated distinctiveness, with no dominant structural dimension and leading communities C7, C9 and C15. Its comparatively balanced profile is more stable than HYBE and slightly less stable than SM. The greatest catalogue concentration belongs to HYBE, whereas SM's descriptive emphasis is integration. These labels synthesise relative evidence without establishing complete multivariate separation.

\subsection{Framework Validation}
\label{sec:framework_validation}

\begin{figure*}[t]
    \centering
    \includegraphics[width=\textwidth]{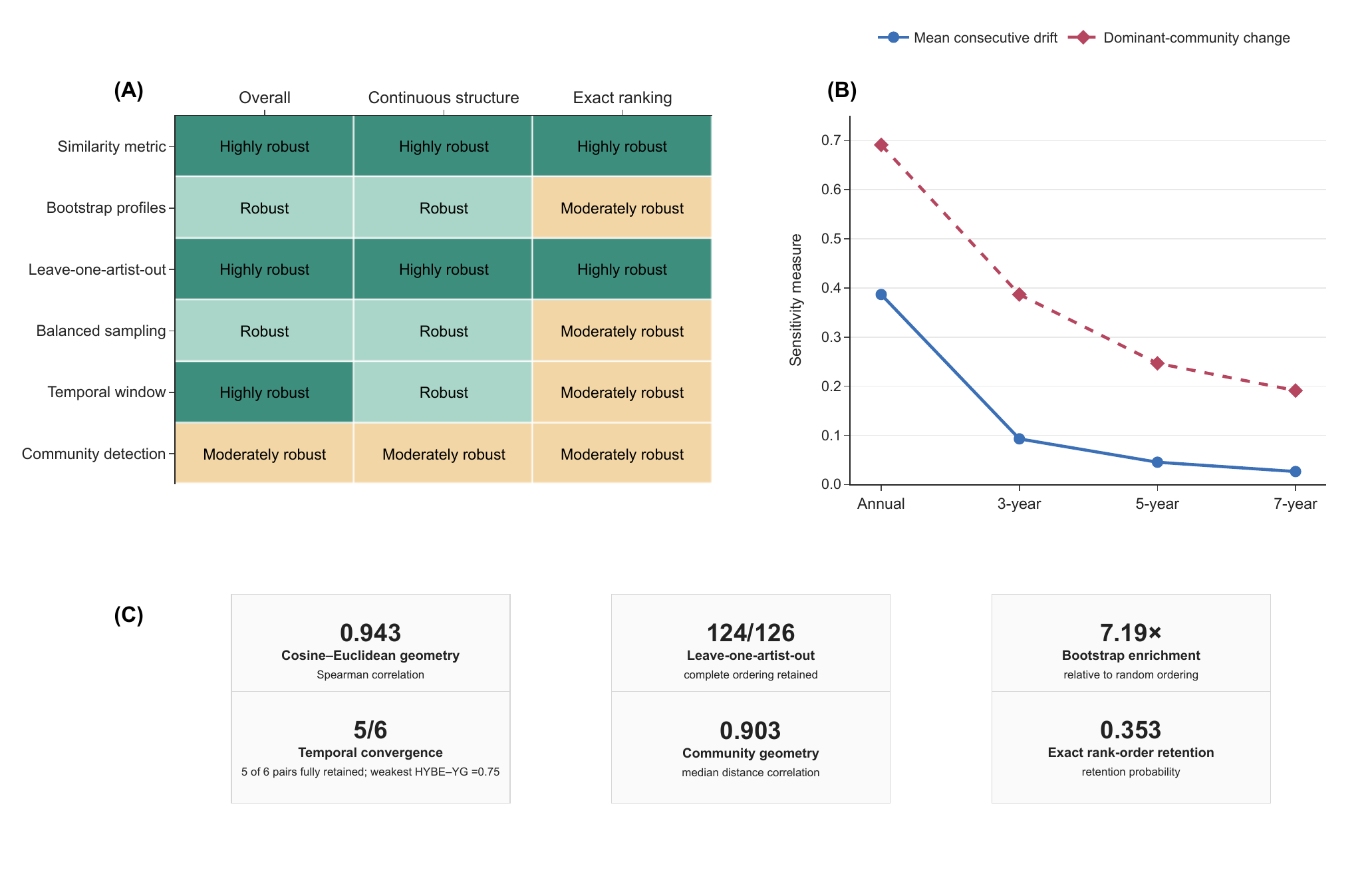}
    \caption{Robustness and sensitivity analysis of the proposed organisational semantic identity framework. (A) Summary of robustness conclusions across the six sensitivity analyses, distinguishing continuous semantic structure from exact ordinal rankings. (B) Sensitivity of temporal semantic drift estimates to alternative temporal aggregation windows. (C) Quantitative robustness evidence derived from similarity-metric sensitivity, bootstrap resampling, leave-one-artist-out analysis, temporal aggregation and community-detection sensitivity.}
    \label{fig:robustness_sensitivity}
\end{figure*}

Figure~\ref{fig:robustness_sensitivity} shows that the principal organisational conclusions are substantially more stable at the level of continuous semantic structure than at the level of exact ordinal ranking. Alternative similarity metrics preserve the company geometry strongly, with a cosine--Euclidean Spearman correlation of 0.943. Leave-one-artist-out analysis retains the complete ordering in 124 of 126 exclusions, demonstrating that the company-level structure is not driven by individual artists. Bootstrap reconstruction yields a 7.19-fold enrichment relative to random ordering, while balanced sampling preserves the broad organisational structure despite unequal catalogue sizes.

Temporal-window sensitivity shows that wider windows progressively reduce short-term drift and dominant-community changes while retaining the broader directional conclusions. Five of the six company pairs preserve full directional convergence across the tested windows, with HYBE--YG providing the weakest retention at 0.75. Community-detection sensitivity similarly preserves continuous geometry, with a median distance correlation of 0.903, although exact rank-order retention is lower at 0.353. These results support the interpretation that broad semantic configurations are robust, while precise ordinal positions should be treated more cautiously.

\begin{figure*}[t]
    \centering
    \includegraphics[width=\textwidth]{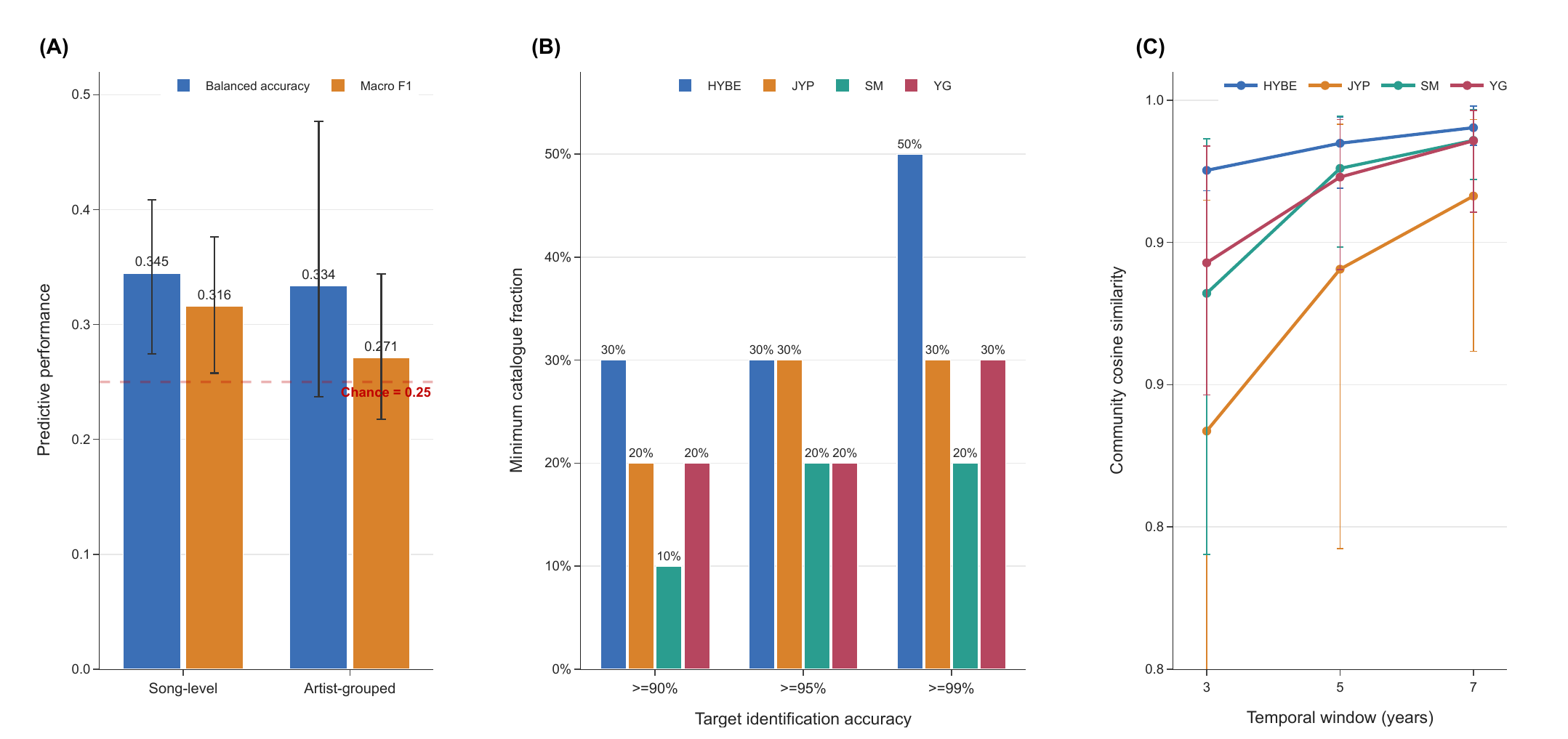}
    \caption{Complementary predictive assessment of the semantic representations. (A) Company prediction performance under song-level and artist-grouped validation protocols. (B) Minimum catalogue fraction required to recover each organisation with predefined identification accuracies. (C) Temporal persistence of organisational semantic communities across alternative rolling-window definitions measured using community cosine similarity.}
    \label{fig:predictive_validation}
\end{figure*}

Figure~\ref{fig:predictive_validation}(A) shows that the fingerprints retain company-specific information under both validation protocols. Song-level balanced accuracy is 0.345 and artist-grouped balanced accuracy is 0.334, compared with the four-class chance baseline of 0.25; the corresponding macro-F1 scores are 0.316 and 0.271. The artist-grouped result is intentionally more conservative because all songs by the same artist remain within a common validation group. The point estimates exceed chance but are modest; they provide complementary evidence of affiliation-related information, not primary validation of semantic identity or near-complete company separability.

Catalogue recovery asks a different question about aggregated evidence within the observed corpus; it is not stronger external validation of the identity construct. At 10\% of the available catalogue, identification accuracy ranges from 0.720 to 0.932 across companies; at 50\%, accuracy reaches at least 0.996 for every company, and at 90\% all four companies are recovered perfectly. Figure~\ref{fig:predictive_validation}(C) further shows high community-level temporal persistence across three-, five- and seven-year windows. These experiments show that partial samples increasingly reproduce the reference fingerprints of their known portfolios, while community composition remains persistent across alternative temporal definitions.

The computational reproducibility audit covered 27 retained analytical scripts spanning the organisation-level workflow. All audited scripts executed successfully, with stochastic components controlled and no reported execution failures. Evidence traceability was assessed by linking each integrated organisational interpretation to the corresponding fingerprint dimensions, temporal qualifications and validation outputs. The validity assessment bounds the supported conclusions to comparative semantic behaviour within the analysed corpus; these conclusions do not establish managerial intent, causal organisational culture or universal organisational characteristics.

\begin{table*}[t]
\centering
\caption{Integrated validation synthesis of the proposed organisational semantic identity framework. The table follows the six validation perspectives introduced in Section~\ref{sec:validation} and concludes with their integrated framework-level assessment.}
\label{tab:validation_synthesis}

\small
\setlength{\tabcolsep}{5pt}

\begin{tabularx}{\textwidth}{p{3.0cm} p{2.5cm} X p{1.7cm}}
\toprule
\textbf{Validation dimension} &
\textbf{Evidence} &
\textbf{Principal finding} &
\textbf{Outcome} \\
\midrule

Statistical validation
&
Artist-level non-parametric inference
&
Company differences are statistically supported for all three novelty-related measures, while the tested diversity and cross-community connectivity differences do not remain significant after Holm correction.
&
Partially supported
\\

Robustness and sensitivity
&
Six predefined perturbation analyses
&
Continuous company geometry remains stable under alternative metrics, resampling, artist exclusion, balanced sampling, temporal windows and community partitions; exact rankings are less stable.
&
Robust structure
\\

Predictive validation
&
Classification, catalogue recovery and temporal persistence
&
Company prediction is above chance under song-level and artist-grouped protocols; partial catalogues become highly identifiable and community composition remains temporally persistent.
&
Informative
\\

Computational reproducibility
&
Audit of 27 retained analytical scripts
&
All audited scripts executed successfully with controlled stochastic components and no reported execution failures.
&
Passed
\\

Evidence traceability
&
Fingerprint, temporal and validation evidence records
&
Every integrated organisational interpretation is linked to explicit quantitative evidence generated by the analytical workflow.
&
Complete
\\

Validity and scope assessment
&
Statistical, computational, data and construct validity
&
The evidence supports comparative, corpus-dependent semantic interpretations; causal and universal organisational claims remain outside the validated scope.
&
Supported within scope
\\

Integrated framework validation
&
Convergence across all six perspectives
&
Together, these checks support within-corpus reproducibility and traceability, while leaving ranking sensitivity, modest predictive performance and cross-domain generality as important qualifications.
&
Supported within the evaluated scope
\\

\bottomrule
\end{tabularx}
\end{table*}

Table~\ref{tab:validation_synthesis} summarises reported within-corpus checks. Its predictive category is complementary, its reproducibility outcome is the original audit, and convergence does not establish independent construct validity or cross-domain generality.

\section{Discussion}
\label{sec:discussion}
\subsection{Collective Differentiation and Contributor Heterogeneity}
\label{sec:discussion_semantic_behaviour}
The central finding is that organisational portfolios differ selectively despite drawing on a shared semantic landscape. Novelty-related measures provide the clearest artist-level inferential evidence, whereas the tested diversity and connectivity measures, together with most community enrichments, do not support differentiation after correction. A portfolio can therefore display recognisable aggregate tendencies without possessing exclusive themes or representing every contributor uniformly. These non-significant results limit claims of separation, although they do not establish equivalence.

Pooled catalogue profiles describe collective output, while artist-level comparisons test whether apparent differences persist across contributors. Reading them together prevents aggregate patterns from obscuring heterogeneity. Leave-one-artist-out and balanced-sampling checks reduce particular sampling concerns, but cannot isolate the effect of affiliation from artist selection, catalogue composition or production practices. Moreover, the static company lookup means that temporal trajectories should not be read as verified histories of corporate ownership.

Artist-grouped balanced accuracy of 0.334, against a chance baseline of 0.25, offers modest complementary evidence that the representation retains affiliation-related information. Because this evaluation uses a fixed corpus representation and catalogue recovery relies on overlapping samples and references, neither constitutes independent measurement of organisational identity. As in computational studies of organisational communication, participant roles and grouping boundaries matter \cite{Swalve2025PartyCommunication}; affiliated lyrics are not official company statements.

\subsection{Continuity and Change}
The temporal profiles distinguish several forms of continuity and change. SM and YG are comparatively stable, JYP combines intermediate composite stability with substantial typical drift, and HYBE shows the greatest relative instability and variability despite declining late-period drift. These differences matter because composite stability, median drift and maximum displacement capture distinct aspects of a trajectory. Identity theory motivates attention to continuity alongside adaptation \cite{Gioia2000AdaptiveInstability}, but these measures do not observe members' self-understanding.

Temporal interpretation is constrained by overlapping windows, uneven observation histories and smoothing that can conceal short transitions. Early, middle and late phases are relative stages within each portfolio's history, not matched calendar periods. Four-company normalisation supplies no absolute stability threshold. Persistence in adjacent windows does not demonstrate independent forecasting or establish historical causes.

\subsection{Computational Contribution and Interpretive Status}
\label{sec:discussion_semantic_identity}
\label{sec:discussion_positioning}
\label{sec:discussion_methodological_contributions}
The methodological contribution lies in bringing established measures together around an inherited semantic foundation \cite{karakucs2026semantic}. The representation locates observations in semantic space, fingerprints describe measurable portfolio properties, and semantic identity provides a qualified synthesis rather than an additional directly observed variable. This complements company representations intended for similarity or prediction \cite{Ito2020CompanyEmbeddings,Gerling2024Company2Vec}, while differing from predefined culture measurements \cite{Li2021CorporateCulture,Koch2023CultureBERT}. Interpretable computational analysis gains value by connecting measurements to a specific empirical question \cite{Su2024HumanInsights}.

Evaluation layers support different claims: tests localise differentiation, robustness assesses sensitivity, prediction assesses retained signal, and reproducibility and traceability make the computational chain inspectable. Their shared foundation prevents treating agreement as independent external validation. Transparency about interpretation is a contribution, not a solution to the construct-validity problem \cite{Joung2026Measurement}.

\subsection{Scope, Limitations and Reuse}
\label{sec:discussion_broader_implications}
\label{sec:discussion_limitations}
\label{sec:discussion_future}
The study covers four company-associated portfolios within a single cultural domain, so its findings cannot establish wider empirical generality. Lyrics are the sole semantic evidence: they do not directly reveal employee values, corporate culture, management or strategy. Nor does affiliation establish who shaped lyrical decisions, making it impossible here to separate organisational influence from contributor selection and portfolio composition.

Fingerprints depend on the inherited corpus, encoder and community partition. Continuous geometry is more robust than exact rankings, and within-corpus reproducibility does not remove representational dependence. Novelty is relative to the semantic representation, not a judgement of artistic originality or managerial innovation. Integrated labels remain comparative descriptions whose correspondence to broader constructs is unverified.

Reuse in institutional publications, political communication or collaborative archives would require defensible contributor and affiliation definitions, appropriate temporal resolution and renewed semantic and construct assessment. Communities and labels cannot simply be transported. Future work could compare encoders and modalities, distinguish within-contributor change from turnover, and evaluate temporally separated forecasts. These require additional data or analyses; they are not findings of the present study.

\section{Conclusion}
\label{sec:conclusion}

This longitudinal comparison of four K-pop company-associated portfolios shows that collectively recognisable semantic patterns need not imply clear separation between organisations. Artist-level evidence of differentiation is concentrated in novelty-related measures; the tested diversity and connectivity measures do not remain significant after correction, and only a small number of community enrichments are supported. Differences in temporal trajectories add another dimension to the static profiles, while the shared semantic landscape and variation among contributors remain essential to their interpretation.

The proposed framework brings these observations together through interpretable fingerprints and a qualified synthesis termed \emph{organisation-level semantic identity}. This corpus-dependent construct describes recurring and evolving characteristics of affiliated textual output; it is not a direct measure of corporate identity, organisational culture or managerial intention. Its contribution is to connect established quantitative measures to a clearly bounded question of collective differentiation and change.

Robustness, catalogue recovery and temporal persistence provide bounded evidence about the measured representations. Modest artist-grouped prediction is complementary evidence of affiliation-related signal, not primary validation of identity. The study remains descriptive, limited to four organisations in one domain, and dependent on its inherited semantic foundation. Cross-domain generality requires further evaluation. Within these limits, the analysis offers a reproducible approach to studying selective collective differentiation and longitudinal change in organisationally associated text.

\section*{Acknowledgements}
The authors acknowledge the use of AI-assisted tools for manuscript preparation, as described below.

\section*{Author Contributions}
Author contributions are provided on the separate title page to preserve anonymous review.

\section*{Statements and Declarations}
\subsection*{Ethical considerations}
This study involved the secondary analysis of existing song lyrics and associated metadata. No human participants were recruited, and no interviews, surveys, or human-participant experiments were conducted. Ethical approval was not required for this research, and no formal ethics application was submitted.

\subsection*{Consent to participate}
Not applicable. The study did not involve human participants.

\subsection*{Consent for publication}
Not applicable. No participant consent for publication was required.

\subsection*{Declaration of conflicting interest}
The authors declared no potential conflicts of interest with respect to the research, authorship, and/or publication of this article.

\subsection*{Funding statement}
The authors received no financial support for the research, authorship, and/or publication of this article.

\subsection*{Data availability}
The study uses existing K-pop lyrics and associated metadata. Copyrighted lyrics and restricted data will not be redistributed. No public release of the full dataset is promised.

\subsection*{Code availability}
A limited public release of code and documentation is planned, pending preparation and rights review. A public code repository has not yet been designated. The planned release will not distribute copyrighted lyrics or restricted data and will not provide full end-to-end reproducibility.

\subsection*{AI-use disclosure}
AI-assisted tools were used to support scientific discussion, language editing and manuscript organisation. The authors reviewed the outputs and take full responsibility for the analyses, interpretations and final manuscript.

\bibliographystyle{SageH}
\bibliography{references_final}
\end{document}